\def\year{2023}\relax
\documentclass[letterpaper]{article} 
\usepackage{flushend}
\usepackage{aaai23}  
\usepackage{times}  
\usepackage{helvet}  
\usepackage{courier}  
\usepackage[hyphens]{url}  
\usepackage{graphicx} 
\def\UrlFont{\rm}  
\usepackage{natbib}  
\usepackage{caption} 
\DeclareCaptionStyle{ruled}{labelfont=normalfont,labelsep=colon,strut=off} 
\usepackage[switch]{lineno}

\usepackage{flushend}
\usepackage{algorithm}
\usepackage{algorithmic}
\usepackage{graphicx}
\usepackage{amsmath}
\usepackage{amssymb}
\usepackage{subfigure}
\usepackage{bbding}
\usepackage{dsfont}
\usepackage{newfloat}
\usepackage{listings}
\floatstyle{ruled}
\newfloat{listing}{tb}{lst}{}
\floatname{listing}{Listing}
\usepackage[switch]{lineno}

\usepackage{wrapfig}
\usepackage{graphicx}
\usepackage{amsmath}
\usepackage{amssymb}
\usepackage{booktabs}
\usepackage{flushend}
\usepackage[dvipsnames]{xcolor}
\usepackage{bm}
\usepackage{multirow}
\usepackage{makecell}
\usepackage{tikz}
\usepackage{comment}
\usepackage{amsmath,amssymb} 

\usepackage{color}
\usepackage[utf8]{inputenc} 
\usepackage[T1]{fontenc}    
\usepackage{url}            
\usepackage{booktabs}       
\usepackage{amsfonts}       
\usepackage{nicefrac}       
\usepackage{microtype}      
\usepackage{xcolor}    

\usepackage{bbding}
\usepackage[capitalize]{cleveref}
\usepackage[misc]{ifsym}
\crefname{section}{Sec.}{Secs.}
\Crefname{section}{Section}{Sections}
\Crefname{table}{Table}{Tables}
\crefname{table}{Tab.}{Tabs.}

\ExplSyntaxOn
\newcommand\latinabbrev[1]{
	\peek_meaning:NTF . {
		#1\@}%
	{ \peek_catcode:NTF a {
			#1.\@ }%
		{#1.\@}}} 
\ExplSyntaxOff

\title{Panoramic Video Salient Object Detection with Ambisonic Audio Guidance} 
\author{
   Xiang Li \textsuperscript{\rm 1}\thanks{This work was done when Xiang Li was an intern at Bytedance.},
   Haoyuan Cao \textsuperscript{\rm 2},
   Shijie Zhao \textsuperscript{\rm 3 \Letter},
   Junlin Li \textsuperscript{\rm 2},
   Li Zhang \textsuperscript{\rm 2},
   Bhiksha Raj \textsuperscript{\rm 1},
}

\affiliations{
    \textsuperscript{\rm 1} Department of Electrical and Computer Engineering, Carnegie Mellon University, PA, USA.\\ 
    \textsuperscript{\rm 2} Bytedance Inc., San Diego, CA, USA.
    \textsuperscript{\rm 3} Bytedance Inc., Shenzhen, China.\\
    \{xl6, bhiksha\}@andrew.cmu.edu, \{haoyuan.cao, zhaoshijie.0526, lijunlin.li, lizhang.idm\}@bytedance.com
}

\usepackage{bibentry}
\begin{document}
\maketitle

\begin{abstract}
Video salient object detection (VSOD), as a fundamental computer vision problem, has been extensively discussed in the last decade. However, all existing works focus on addressing the VSOD problem in 2D scenarios. 
With the rapid development of VR devices, panoramic videos have been a promising alternative to 2D videos to provide immersive feelings of the real world. 
In this paper, we aim to tackle the video salient object detection problem for panoramic videos, with their corresponding ambisonic audios. 
A multimodal fusion module equipped with two pseudo-siamese audio-visual context fusion (ACF) blocks is proposed to effectively conduct audio-visual interaction. 
The ACF block equipped with spherical positional encoding enables the fusion in the 3D context to capture the spatial correspondence between pixels and sound sources from the equirectangular frames and ambisonic audios. 
Experimental results verify the effectiveness of our proposed components and demonstrate that our method achieves state-of-the-art performance on the ASOD60K dataset.
\end{abstract}
\section{Introduction}
Video salient object detection (VSOD) aims to find the most visually distinctive objects in a video. VSOD for 2D videos has been attracting considerable attention \cite{su2022unified,liu2021swinnet,ren2021unifying} due to its wide applications in real-world scenarios, such as video editing and video compression. While, for panoramic videos which have a very different format and viewing environment, how to effectively detect salient objects is still an open problem. Since human attention is usually influenced by acoustic signatures that are naturally synchronized with visual objects in audio-bearing video recordings, some VSOD methods for 2D videos \cite{tsiami2020stavis,cheng2021audio} introduce acoustic modality to facilitate the saliency discrimination. Different from mono or binaural audio used in 2D videos, ambisonic audio is utilized to create
immersive feelings of the real world in panoramic videos. In this work, we focus on how to detect the salient objects in panoramic videos with their corresponding ambisonic audios. To the best of our knowledge, we are the first to tackle the VSOD problem for panoramic scenarios.

\begin{figure}[t!]
    \centering
    \includegraphics[width=\linewidth]{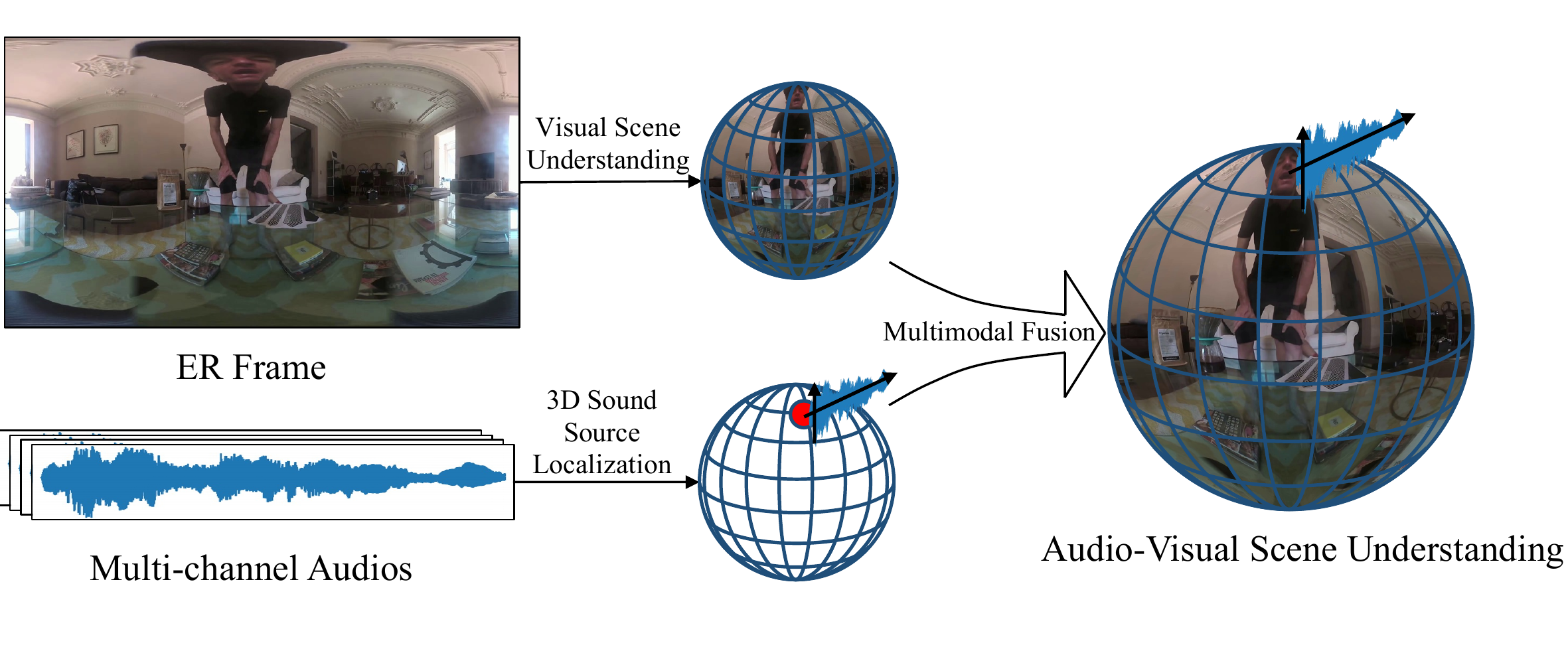}
    \vspace{-0.7cm}
    \caption{We study the problem of how to utilize the ambisonic audio to facilitate the panoramic video salient object detection. Since ambisonic audio contains rich spatial information, we can directly localize the sound sources and then fuse them with visual cues. On the other hand, since the ER frame has distortions and cannot reflect the 3D position of pixels, projecting the pixels back to 3D space is essential for effective multimodal fusion.
    }
    \label{fig:teaser}
\end{figure}

The panoramic video contains omnidirectional contexts and represents each pixel on a 3D sphere. Unlike 2D videos that display with a stable viewpoint and fixed environmental audios, panoramic videos are typically supported by VR headsets which provide a head direction-adaptive field of view (FOV) and ambisonic audios. With multi-channel audio recordings, ambisonic audios encode the 3D location of the sound source and the VR headset can further adjust the original audio to provide a real sense of sound source to the user given the movement of the head. Previous works \cite{tsiami2020stavis,cheng2021audio} have demonstrated the audio shows non-trivial influence on human attention in 2D videos. For VSOD in panoramic videos, as shown in Figure~\ref{fig:teaser}, since the ambisonic audio can directly reflect the 3D sound source location, we consider it should perform a more important role compared to 2D scenarios. 

On the other hand, previous panoramic video processing approaches cannot preserve the spatial relationship of panoramic videos. Due to the unique format of panoramic videos, it is difficult to store or transmit the raw video using current video coding techniques. Therefore, equirectangular (ER) projection is commonly leveraged to transform the panoramic video into a regular 2D format. However, the ER projection will involve not only a separation along the longitude of the sphere but also distortions in the polar area, which severely barrier the effective VSOD. Previous methods address the polar distortions by introducing a cube projection \cite{cheng2018cube} which projects the sphere to a cube and expands each face to ease the distortions. However, the padded cube map severely destroys the spatial relationship between each face which obstacles the global understanding of the frame.

In this paper, we purpose a framework for audio-visual salient object detection in panoramic scenarios. Considering the rich spatial and semantic information encoded in the ambisonic audios, we first use a pretrained acoustic encoder to extract location and semantic embeddings of 3D sound sources, and then introduce audio-visual context fusion (ACF) blocks to enhance the visual features by acoustic cues. 
Ideally, the 3D location of sound sources should be utilized as ground-truth to finetune the acoustic network while it is very difficult to obtain for common panoramic videos.
To tackle this problem, inspired by label-guided distillation \cite{zhang2022lgd}, 
the ground-truth label of salient object is employed in the multimodal fusion by enforcing consistency between teacher (equipped with label) and student branch. By learning better audio-visual correspondence, we can transfer the spatial information encoded in the visual objects to supervise acoustic modality. In particular, to reflect the true 3D location of objects and mitigate the influence of distortions in ER frames, we map the 2D coordinates of pixels back to the 3D sphere and encode them in the positional encoding during the fusion. In this way, the model can capture the true spatial position of each pixel. Our contributions are summarized as:
\begin{itemize}
    \item We propose an audio-visual video salient object detection framework for panoramic scenarios. To the best of our knowledge, we are the first to tackle this problem.
    \item We introduce a label-guided audio-visual fusion module to effectively utilize the rich spatial and semantic information encoded in the ambisonic audio recordings synchronized with panoramic videos.
    \item Our model achieves state-of-the-art results on the ASOD60K dataset. Extensive experiments are conducted to illustrate the effectiveness of our method.
\end{itemize}

\section{Related Works}
\subsection{Video Salient Object Detection}
VSOD aims to find the most visually salient objects in the video. Conventional methods usually leverage color contrast \cite{achanta2009frequency}, motion prior \cite{zhang2019self}, background prior \cite{yang2013saliency} and center prior \cite{jiang2013submodular,bertasius2017unsupervised} to distinguish the salient regions. However, most of those methods are limited by the low representative ability of hand-crafted features. Recent methods leverage deep-learning approaches to tackle the VSOD problem. To utilize the temporal information, FCNS \cite{wang2017video} first leverages FCN for static saliency prediction and then post-process them by another dynamic FCN. Similarly, DSRFCN3D \cite{le2017deeply} introduces 3D convolution for temporal aggregation. Optical flow reflecting the motion also serves as a strong cue for VSOD task in \cite{li2018flow,li2019motion}. Recently, more advanced structures are leveraged to better understand the spatio-temporal correspondence. For example, ConvLSTM \cite{li2018flow,song2018pyramid,fan2019shifting} is adopted to construct long-term temporal relation. With the strong ability of transformer \cite{vaswani2017attention} to model the complex relationship, it has achieved promising result in VSOD \cite{liu2021swinnet,ren2021unifying}. 

\subsection{Panoramic Saliency Detection}
Saliency detection aims to predict the region of human eye fixation in the video. For image-level saliency prediction \cite{cheng2018cube,suzuki2018saliency},\cite{cheng2018cube} propose a cube padding operation to project the panoramic frame to a cube with fewer distortions on each face. \cite{zhu2018prediction} first map the ER frame to a sphere then predict the saliency from each viewport. For video-level saliency prediction \cite{cheng2018cube,nguyen2018your}, Nguyen et al. \cite{nguyen2018your,zhang2018saliency} proposes a method that leverages CNN and LSTM for saliency prediction. In \cite{zhang2018saliency}, spherical CNN is introduced to directly handle panoramic videos. In addition, audio is introduced in panoramic saliency prediction \cite{chao2020audio} given its strong ability to influence human attention.

\subsection{Video Object Segmentation}
Video object segmentation (VOS) can be categorized as unsupervised \cite{wang2019learning,ren2021reciprocal}, semi-supervised \cite{wang2021swiftnet} and referring \cite{wu2022language,li2022r} VOS. The most relevant type to this work is the unsupervised VOS (UVOS) which aims to segment primary object regions from the background in videos. Early methods tackle the UVOS problem by object proposal \cite{kim2002fast}, temporal trajectory \cite{fragkiadaki2012video} and saliency prior \cite{wang2015robust,wang2015saliency}. More recently, deep learning-based methods are proposed for modeling the spatio-temporal information. 
MATNet \cite{zhou2020matnet} uses a motion-attentive transition to model motion cues and spatio-temporal representation. RTN \cite{ren2021reciprocal} leverages long-range intra-frame contrast, temporal coherence, and motion-appearance similarity to enhance the appearance feature representation. In addition to VOS, video instance segmentation \cite{li2022hybrid,li2022video,li2022online} is also relevant to this work. Recently, some works extend the video segmentation task to multimodal by considering audio \cite{zhou2022audio} or signals \cite{zhao2022self,huang2021towards,huang2021forgery}.
\section{Method}
\begin{figure*}[t!]
    \centering
    \includegraphics[width=\linewidth]{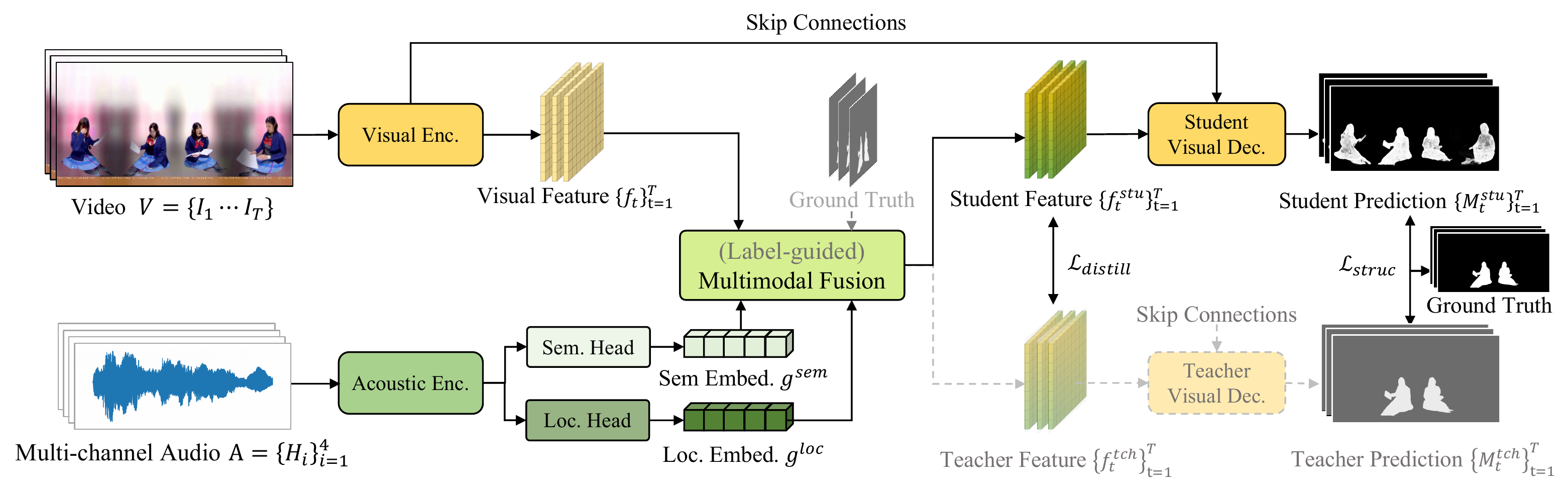}
    \caption{\textbf{Pipeline Overview.} We use separate encoders to extract multimodal features. For a video clip $V=\{I_1\cdots I_T\}$, a visual encoder is utilized to extract visual feature $\{f_t\}_{t=1}^T$. For audio input $A=\{H_i\}_{i=1}^4$, a two-brunch acoustic encoder is employed to extract the semantic embedding $g^{sem}$ and location embedding $g^{loc}$. After that, a label-guided multimodal fusion module is introduced to effectively fuse the multimodal features, which outputs a student feature $\{f_t^{stu}\}_{t=1}^T$ and a teacher feature $\{f_t^{tch}\}_{t=1}^T$. Two decoders are leveraged to decode the final predictions $\{M_t^{stu}\}_{t=1}^T$ and $\{M_t^{tch}\}_{t=1}^T$ from compacted features $\{f_t^{stu}\}_{t=1}^T$ and $\{f_t^{tch}\}_{t=1}^T$ respectively. In particular, to enhance multimodal communication, a distillation loss $\mathcal{L}_{ditill}$ is adopted between student feature $\{f_t^{stu}\}_{t=1}^T$ and (label-guided) teacher feature $\{f_t^{tch}\}_{t=1}^T$. A structure loss $\mathcal{L}_{struc}$ is utilized as the objective. The {\bf\color{gray}{gray}} color indicates components that are only used during training.
    }
    \label{fig:pipeline}
\end{figure*}

\subsection{Overview}
Given a video clip $V=\{I_t\}_{t=1}^T$ of $T$ frames and its corresponding multi-channel audio recordings $A=\{H_i\}_{i=1}^4$, we predict the salient object $\{M_{t}\}_{t=1}^T$ effectively with our method. The method overview is illustrated in Figure~\ref{fig:pipeline}. The pipeline can be boiled down into three parts: acoustic and visual encoders, a label-guided multimodal fusion module, and decoders. We first leverage a visual and an acoustic encoder to extract visual $\{f_t\}_{t=1}^T$ and acoustic features $g^{sem}$, $g^{loc}$. The acoustic features contains a semantic embedding $g^{sem}$ and a location embedding $g^{loc}$ which encodes the category and 3D location of sound sources respectively. After that, a multimodal fusion module is utilized to enable audio-visual interaction. Specifically, the multimodal fusion module contains two pseudo-siamese blocks - a student block that fuses audio-visual information using multimodal attention and a teacher block that shares the same structure of the student block while taking an additional ground truth as input to guide the fusion. The output student features $\{f_t^{stu}\}_{t=1}^T$ and (label-guided) teacher features $\{f_t^{tch}\}_{t=1}^T$ are sent to visual decoders equipped with skip connections to generate the final salient object predictions $\{M_t^{stu}\}_{t=1}^T$ and $\{M_t^{tch}\}_{t=1}^T$. A distillation loss between student feature $\{f_t^{stu}\}_{t=1}^T$ and (label-guided) teacher features $\{f_t^{tch}\}_{t=1}^T$ is adopted to help the multimodal interaction and a structure loss between prediction $\hat{M}_t$ and ground truth $M_t$ is used as the task objective for salient object detection. 

\subsection{Label-guided Multimodal Fusion}
The label-guided multimodal fusion module contains two pseudo-siamese audio-visual context fusion (ACF) blocks equipped with spherical positional encoding to align the correspondence between 3D sound sources and pixels. The output of student and teacher block are student feature $\{f_t^{stu}\}_{t=1}^T$ and teacher feature $\{f_t^{tch}\}_{t=1}^T$ respectively.

\subsubsection{Spherical positional encoding.}

\begin{figure}[htbp!]
    \centering
    \includegraphics[width=0.7\linewidth]{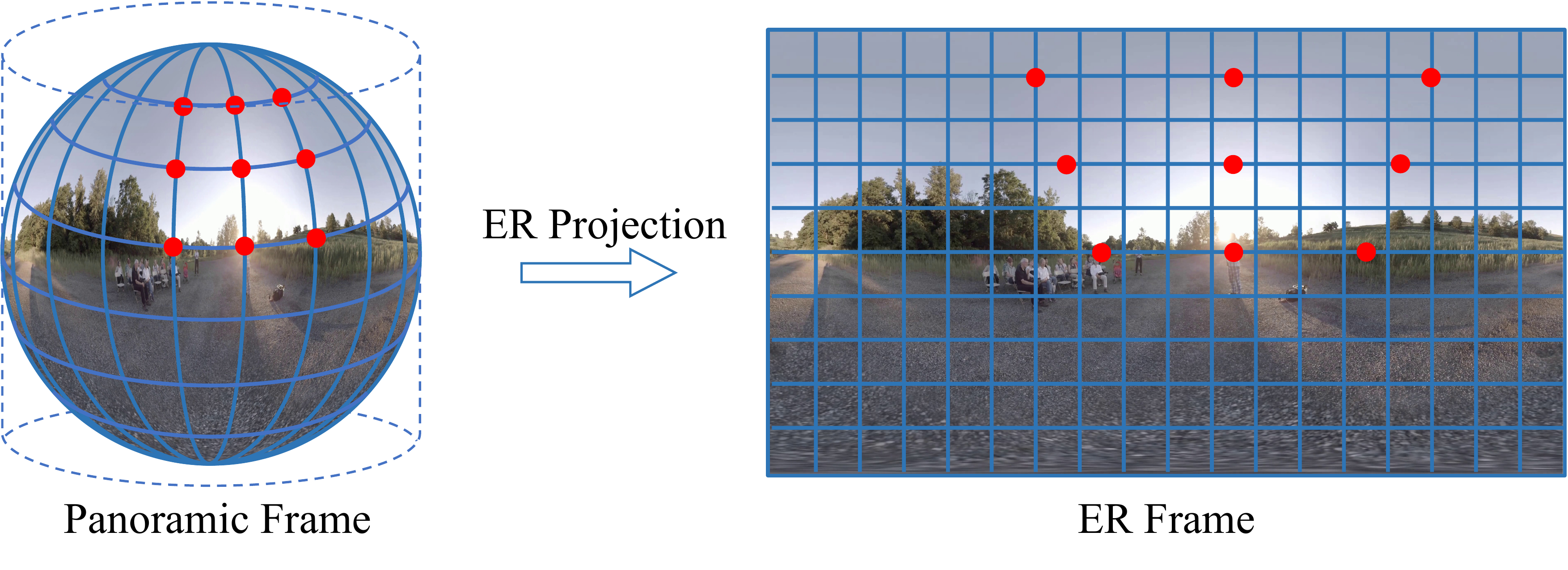}
    \vspace{-0.2cm}
    \caption{\textbf{Illustration of ER Projection.} Severe distortions can be observed in the polar areas.
    }
    \label{fig:erp}
\end{figure}

ER frame is a commonly used format to transmit and store panoramic videos \cite{cai2022overview}. However, as shown in Figure~\ref{fig:erp}, the ER frame suffers from severe distortions in the polar regions. To tackle this problem, we adopt the position-agnostic attention mechanism and propose a spherical positional encoding to compensate for the distortion in the ER frame. Different from regular positional encoding \cite{vaswani2017attention}, spherical positional encoding first re-projects the plane coordinates of each pixel back to the 3D sphere and generates positional encoding based on the 3D coordinates. In this way, each pixel can reflect its true 3D position thus avoiding the severe distortion in the polar region and inevitable separation along a longitude in the ER frame. In particular, the spherical positional encoding can be computed as $\mathrm{PE}(\mathrm{pos_{3D}}, 2i) = \mathrm{sin}(\mathrm{pos_{3D}}/10000^{2i/\frac{d}{3}})$ and $\mathrm{PE}(\mathrm{pos_{3D}}, 2i+1) = \mathrm{cos}(\mathrm{pos_{3D}}/10000^{2i/\frac{d}{3}})$ where $\mathrm{pos_{3D}}$ can be $x$, $y$, $z$ coordinates and $i$ is the dimension. The 2D-to-3D transformation can be computed as
\vspace{-0.2cm}

\begin{equation}
    x = sin\frac{v}{R}cos\frac{u}{R},\,
    y = sin\frac{v}{R}sin\frac{u}{R},\,
    z = cos\frac{v}{R}
\end{equation}
where $(u,v)$ and $(x,y,z)$ are the 2D and 3D coordinate of each pixel respectively. $R=\frac{W}{2\pi}$ where $W$ is the width of the frame. Spherical positional encoding (SPE) is employed to encode spatial information for visual representation during cross-modal attention.

\begin{figure}[t!]
    \centering
    \includegraphics[width=\linewidth]{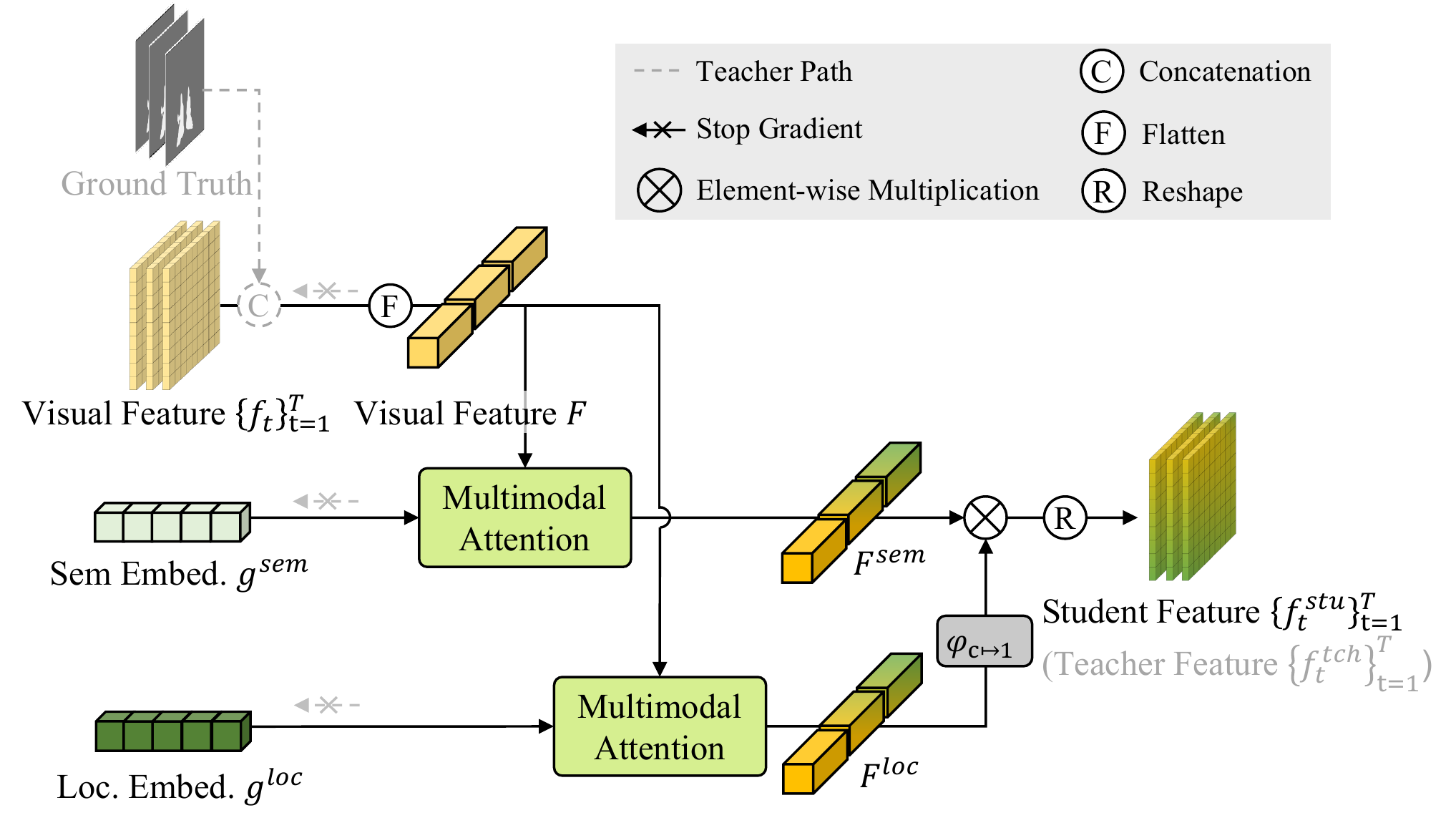}
    \caption{Illustration of ACF block used in the (label-guided) multimodal fusion process. \textbf{Student block}: The visual feature $\{f_t\}_{t=1}^T$ is first flattened and added with positional encoding then conducts multimodal attention with acoustic semantic embedding $g^{sem}$ and location embedding $g^{loc}$ respectively. After that, we form a pixel-wise weighting from fused feature $F^{loc}$ to modulate $F^{sem}$. The output of student block is denoted as $\{f_t^{stu}\}_{t=1}^T$. \textbf{Teacher block}: Different from student block, we concatenate ground-truth mask with the visual feature $\{f_t\}_{t=1}^T$ to help the network learn better representation in the multimodal fusion. Since teacher block is only employed during training, we truncate the gradients before the multimodal fusion in the teacher block. The output of teacher block is denoted as $\{f_t^{tch}\}_{t=1}^T$.
    }
    \label{fig:mm_fusion}
\end{figure}

\subsubsection{Student block.}
As shown in Figure~\ref{fig:mm_fusion} (without the {\color{gray} gray} parts), to fuse the rich information encoded in visual and acoustic features, we utilize multimodal attention to enable audio-visual context interaction. We first concatenate and then flatten the visual feature $\{f_t\}_{t=1}^T$ to form $F=flatten(f_1\oplus\dots\oplus f_T)\in\mathbb{R}^{C\times THW}$. 
After that, spherical and regular positional encodings \cite{vaswani2017attention} are added to visual feature $F$, and acoustic feature $g^{sem}$ and $g^{loc}$, respectively, to help the network capture spatial information. The multimodal attention can be computed by
\vspace{-0.2cm}

\begin{equation}
    h^{aud}=\mathrm{LN}( \mathrm{MCA}(F, g^{aud}) + F)
\end{equation}
\begin{equation}
    F^{aud}=\mathrm{LN}(\mathrm{FFN}(h^{aud}) + h^{aud})
\end{equation}
where $\mathrm{MCA}$, $\mathrm{FFN}$ and $\mathrm{LN}$ are multi-head cross-attention \cite{ye2019cross}, feed-forward network and layer-normalization respectively. In particular, we generate the query $Q$ from $F$ and key $K$, value $V$ from $g^{aud}$ by linear projections. The $\mathrm{MCA}(F, g^{aud})$ can be computed by $\mathrm{Softmax}(\frac{K^\mathrm{T}Q}{\sqrt{d}})V$, where $d$ is the dimension of query $Q$. The student feature $F^{stu}$ can be computed by
\vspace{-0.2cm}

\begin{equation}
    F^{stu} = F^{sem} \odot \varphi_{C\mapsto 1}(F^{loc})
\end{equation}
where $\varphi_{C\mapsto 1}$ denotes a convolution to reduce channel from $C$ to $1$ and $\odot$ denotes element-wise multiplication. The final output is  $\{f_t^{stu}\}_{t=1}^T=\mathrm{Reshape}(F^{stu})$.

\subsubsection{Teacher block.}
The audio encoder is pretrained on a 3D sound source localization dataset \cite{guizzo2021l3das21} while it is difficult to obtain a 3D location of vocal objects in panoramic videos during main training. Inspired by previous work \cite{zhang2022lgd}, we build a pseudo-siamese teacher block to help the network capture accurate spatial information by introducing ground-truth annotation to the multimodal fusion. As shown in Figure~\ref{fig:mm_fusion}  (with the {\color{gray} gray} parts), the additional ground truth is downsampled and concatenated with visual feature $\{f_t\}_{t=1}^T$ in the teacher block before conducting the multimodal interaction. Similar to student block, the final output of teacher block is denoted as $\{f_t^{tch}\}_{t=1}^T$. Note that the teacher block does not share weights with the student block and the gradient of teacher block is truncated to avoid influencing the encoder. 

\subsection{Encoder}
We use separate encoders to extract visual and acoustic features.

\subsubsection{Visual encoder.}

\begin{figure}[h!]
    \centering
    \includegraphics[width=0.9\linewidth]{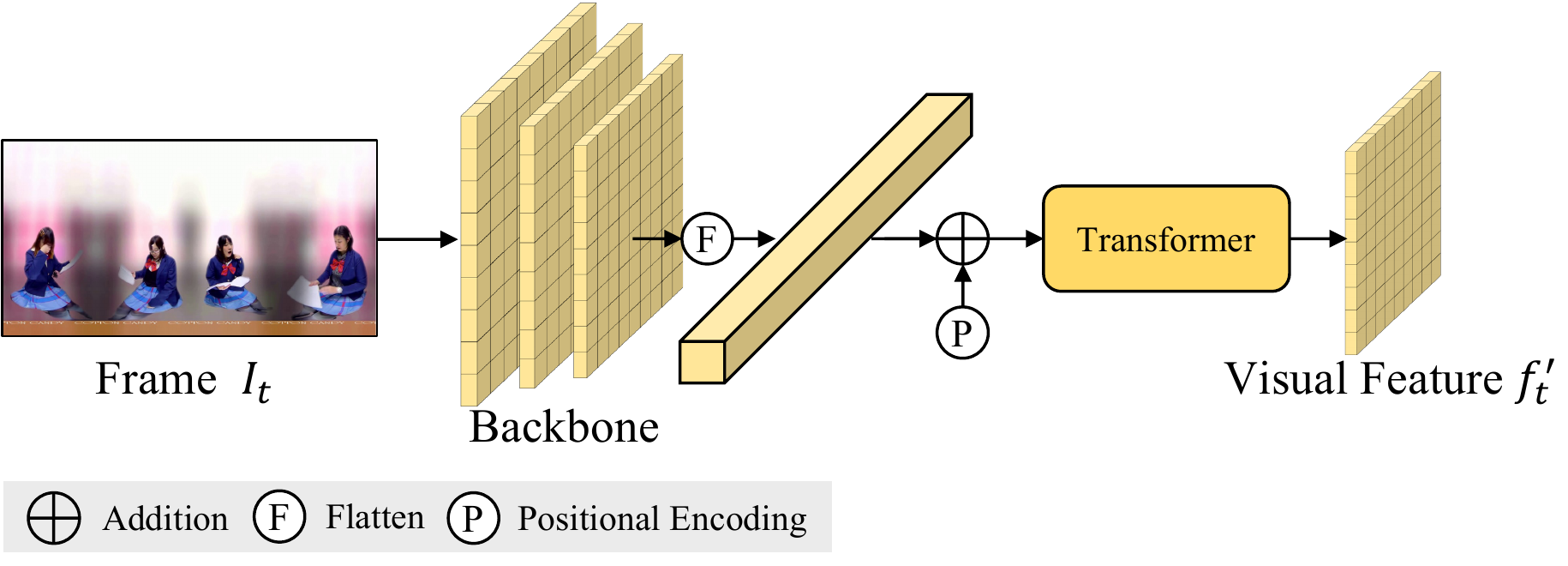}
    \caption{\textbf{Single-frame visual feature extraction.} The ER frame $I_t$ is first processed by a backbone and then the extracted features are fed to a transformer to capture spatial information. The single-frame feature is denoted as $f_t^\prime$.
    }
    \label{fig:enc}
\end{figure}
We first project the panoramic video to 2D frames $\{I_1\cdots I_T\}$ using ER projection and then feed them to the backbone. As shown in Figure~\ref{fig:enc}, a transformer encoder on top of the ResNet-50 \cite{he2016deep} is adopted to mitigate the severe distortion in the ER frame. In addition, a temporal non-local block as \cite{yan2019semi} is also leveraged to enable the temporal interaction on the extracted features $\{f_t^\prime\}_{t=1}^T$ from the backbone. We denote the features after temporal aggregation as $\{f_t\}_{t=1}^T$ where $f_t\in\mathbb{R}^{C\times H\times W}$.

\subsubsection{Acoustic encoder.}
Multi-channel audio contains the 3D location and category information of the sound source which have a great impact on the choosing of the salient objects. 
To extract the acoustic features, we leverage three CNN layers followed by two bi-directional GRU layers as our acoustic encoder and three linear layers for both semantic and location head (detailed structure available in supplementary) \cite{adavanne2018sound}. Since it is difficult to obtain the real-world sound source location in panoramic videos, we first pretrain the acoustic encoder on a 3D sound source localization and sound event classification dataset, L3DAS \cite{guizzo2021l3das21}. We remove final linear layer in each head to form the semantic embedding $g^{sem}\in\mathbb{R}^{C\times L}$ and location embedding $g^{loc}\in\mathbb{R}^{C\times L}$.  

\subsection{Decoder}
We adopt the same structure for decoding $\{f_t^{stu}\}_{t=1}^T$ and $\{f_t^{tch}\}_{t=1}^T$. For decoding the salient object prediction, we follow the FPN structure \cite{lin2017feature} to fuse the low-level features. Let the output salient object prediction be $\{M_{t}^{stu}\}_{t=1}^T\in \mathbb{R}^{H_o\times W_o}$ and $\{M_{t}^{tch}\}_{t=1}^T\in \mathbb{R}^{H_o\times W_o}$ for student and teacher branch respectively, where $H_o$ and $W_o$ are the height and width of the output.

\begin{table*}[t]
\centering
\scalebox{1}{
    \begin{tabular}{l|p{0.5cm}<{\centering}p{0.5cm}<{\centering}p{0.5cm}<{\centering}p{0.5cm}<{\centering}|p{0.5cm}<{\centering}p{0.5cm}<{\centering}p{0.5cm}<{\centering}p{0.5cm}<{\centering}|p{0.5cm}<{\centering}p{0.5cm}<{\centering}p{0.5cm}<{\centering}p{0.5cm}<{\centering}|p{0.5cm}<{\centering}p{0.5cm}<{\centering}p{0.5cm}<{\centering}p{0.5cm}<{\centering}} 
    \hline
    \hline
    \multirow{2}*{Method} &  \multicolumn{4}{c|}{Miscellanea (Test1)} & \multicolumn{4}{c|}{\quad Music (Test2)\quad} & \multicolumn{4}{c|}{\quad Speaking (Test3)\quad} & \multicolumn{4}{c}{\quad ASOD60K-Test All\quad}\\
    \cline{2-5}\cline{6-9}\cline{10-13}\cline{13-17}
    ~ & $F_\beta\uparrow$ & $S_\alpha\uparrow$ & $E_\phi\uparrow$ & $\mathcal{M}\downarrow$ & $F_\beta\uparrow$ & $S_\alpha\uparrow$ & $E_\phi\uparrow$ & $\mathcal{M}\downarrow$ & $F_\beta\uparrow$ & $S_\alpha\uparrow$ & $E_\phi\uparrow$ & $\mathcal{M}\downarrow$ & $F_\beta\uparrow$ & $S_\alpha\uparrow$ & $E_\phi\uparrow$ & $\mathcal{M}\downarrow$\\
    \hline
    \multicolumn{17}{c}{Image-Level Methods} \\
    \hline
    CPD-R & .248 & .654 & .645 & .035 & .272 & .608 & .632 & .018 & .228 & .588 & .657 & .026 & .243 & .609 & .648 & .026 \\
    SCRN & .250 & .665 & .615 & .046 & .341 & .683 & .664 & .023 & .276 & .636 & .642 & .034 & .286 & .655 & .641 & .034 \\
    F3Net & .257 & .655 & .629 & .040 & .358 & .662 & .749 & .021 & .308 & .626 & .692 & .027 & .310 & .642 & .691 & .029 \\
    MINet & .238 & .650 & .625 & .050 & .380 & .670 & .716 & .020 & .261 & .590 & .635 & .053 & .286 & .624 & .652 & .044 \\
    LDF & .280 & .663 & .626 & .044 & .389 & .671 & .753 & .023 & .309 & .625 & .711 & .037 & .322 & .645 & .701 & .035 \\
    CSFR2 & .238 & .652 & \bf.642 & .033 & .347 & .665 & .693 & .018 & .285 & .636 & .700 & .026 & .290 & .646 & .684 & .026 \\
    GateNet & .285 & .677 & .651 & .044 & .290 & .673 & .616 & .018 & .260 & .633 & .638 & .034 & .273 & .653 & .636 & .033 \\
    \hline
    \multicolumn{17}{c}{Video-Level Methods} \\
    \hline
    COSNet & .147 & .610 & .553 & \bf.031 & .220 & .557 & .541 & .016 & .176 & .572 & .570 & \bf.023 & .181 & .582 & .559 & \bf.023 \\
    RCRNet & .272 & .661 & .640 & .034 & .403 & \bf.695 & .738 & .019 & .282 & .632 & .687 & .030 & .310 & .654 & .688 & .029 \\
    PCSA & .123 & .604 & .574 & .034 & .310 & .657 & .645 & .022 & .150 & .571 & .534 & .026 & .184 & .600 & .570 & .027 \\
    3DC-Seg & .300 & .668 & .618 & .062 & .326 & .635 & .632 & .046 & .289 & .629 & .592 & .056 & .300 & .640 & .608 & .055 \\
    RTNet & .240 & .622 & .634 & .038 & .365 & .638 & .766 & .020 & .194 & .555 & .668 & .028 & .247 & .591 & .683 & .025 \\
    \hline
    \textbf{Ours} & \bf.355 & \bf.714 & .617 & .037 & \bf.483 & .682 & \bf.834 & \bf.016 & \bf.382 & \bf.658 & \bf.742 & .024 & \bf.404 & \bf.678 & \bf.732 & .026 \\
    \hline
    \hline
    \end{tabular}
}
\caption{\textbf{Comparison to state-of-the-art salient object detection methods on ASOD60K dataset.} $\uparrow$ means larger is better and $\downarrow$ means smaller is better. \textbf{Bold} means the state-of-the-art performance.}
\label{tab:main results}
\end{table*}

\subsection{Loss Function}
The overall objective of our proposed method is composed of structure losses for student and teacher branch $\mathcal{L}_{struc}^{stu}$, $\mathcal{L}_{struc}^{tch}$ and a distillation loss $\mathcal{L}_{distill}$
\vspace{-0.2cm}

\begin{equation}
    \mathcal{L} = \mathcal{L}_{struc}^{stu} + \mathcal{L}_{struc}^{tch} + \lambda_{distill}\mathcal{L}_{distill}
\end{equation}
where $\lambda_{distill}$ is a scalar to balance the losses.
\subsubsection{Structure loss.}
Following previous method \cite{chen2022video}, we leverage a combination of binary cross entropy loss and Dice loss \cite{milletari2016v} as the objective for salient object detection. 
\vspace{-0.2cm}

\begin{equation}
\mathcal{L}_{struc}=\sum_{t=1}^T\mathcal{L}_{bce}(M_t, \hat{M}_t)+\lambda_{dice}\sum_{t=1}^T\mathcal{L}_{dice}(M_t, \hat{M}_t)    
\end{equation}
where $M_t$ and $\hat{M}_t$ are predicted and ground-truth salient maps respectively. $\lambda_{dice}$ is a scalar. $M_t$ can be $M_t^{stu}$ and $M_t^{tch}$ for computing $\mathcal{L}_{struc}^{stu}$ and $\mathcal{L}_{struc}^{tch}$.

\subsubsection{Distillation loss.}
To help the student block learn the audio-visual correspondence, we enforce a consistency between $f_t^{tch}$ and $f_t^{stu}$. A MSE loss is utilized as the constraint 

\vspace{-0.2cm}
\begin{equation}
\mathcal{L}_{distill}=\sum_{t=1}^T\mathcal{L}_{MSE}(f_t^{tch},f_t^{stu})
\end{equation}

\subsection{Inference}
Since the purpose of introducing teacher block is to facilitate network learn accurate audio-visual correspondence during training, we disable the teacher block and only keep the student block during inference.
\section{Experiment}
\subsection{Dataset and Metrics}
\paragraph{Dataset.}
We conduct experiments on the ASOD60K dataset \cite{zhang2021asod60k} which is an audio-induced salient object detection benchmark for panoramic videos. There are 62,455 frames with 10,465 instance-level ground truths in the dataset. In particular, each video corresponds to a 4-channel ambisonic audio recording. The ground-truth salient objects are determined by the eye fixation of 40 participants who viewed the video with HTC Vive HMD headset. The test set of ASOD60K contains three subsets split by sound event classes - miscellanea, music, and speaking.

\paragraph{Metrics.}
To evaluate the performance of video salient object detection, we employ the adaptive F-Measure $F_{\beta}$ \cite{achanta2009frequency}, adaptive E-Measure $E_{\phi}$ \cite{fan2018enhanced}, S-Measure $S_{\alpha}$ \cite{fan2017structure} and Mean Absolute Error (MAE) $\mathcal{M}$ \cite{borji2015salient}.

\subsection{Implementation Details}
Following the benchmark setting\cite{zhang2021asod60k}, our model is first pre-trained on DUST dataset \cite{wang2017learning} and then finetuned on ASOD60K \cite{zhang2021asod60k}. The model is trained for 20 epochs with a learning rate of 1e-4. We adopt a $\mathrm{batchsize}$ of 2 and an AdamW \cite{loshchilov2017decoupled} optimizer with weight decay $0$. All images are cropped to have the longest side of 832 pixels and the shortest side of 416 pixels during training and evaluation. The window size is set to 3. The $\lambda_{distill}$ is set to 5.0 and $\lambda_{dice}$ is set to 1 if no specification. We leverage a 3-layer transformer encoder \cite{dosovitskiy2020image} on top of the ResNet-50 \cite{he2016deep} to extract visual features. We leverage an augmented SELDNet \cite{adavanne2018sound} as our acoustic encoder. Our method is implemented with PyTorch.

\begin{figure*}[t!]
    \centering
    \includegraphics[width=\linewidth]{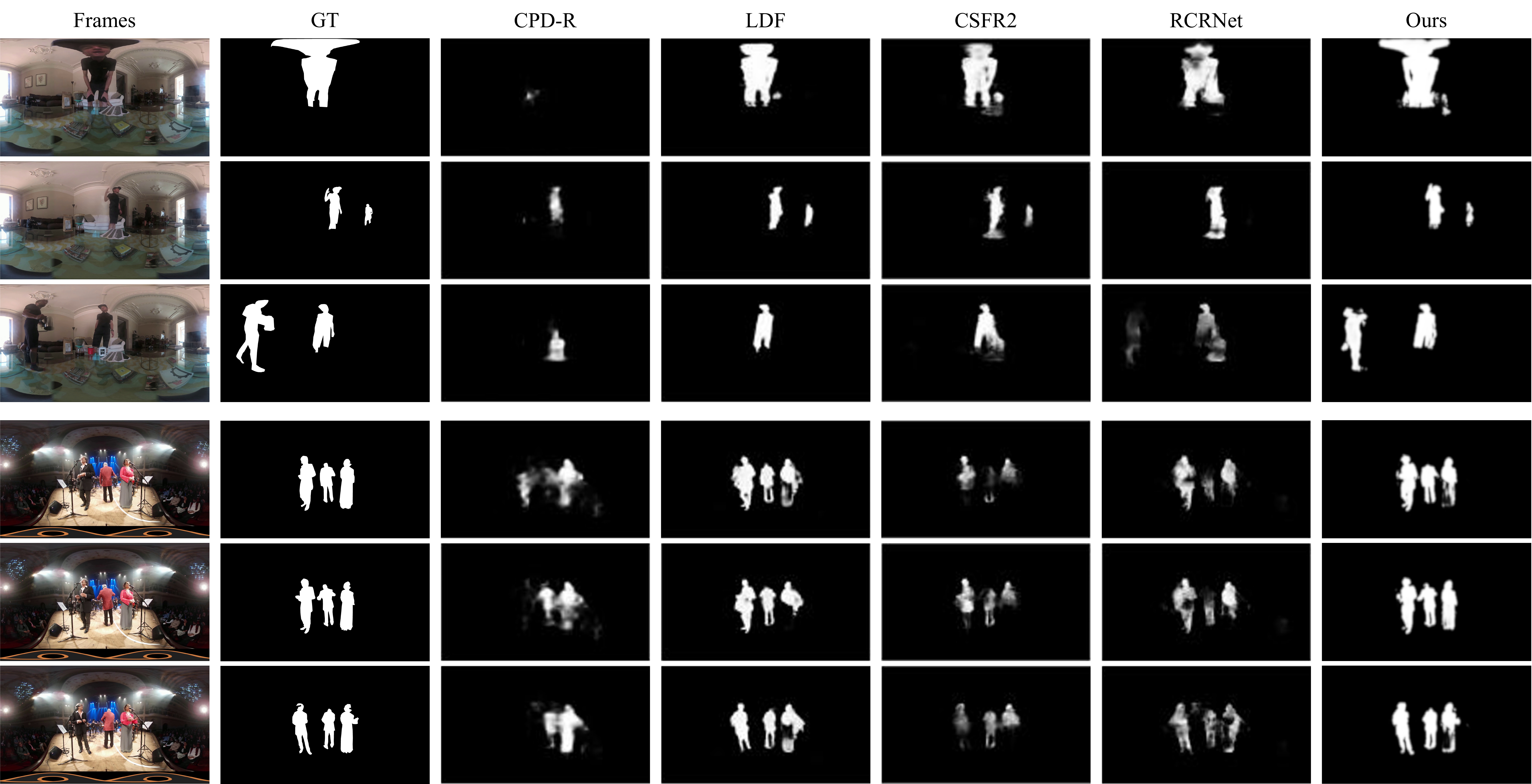}
    \caption{Qualitative comparison to state-of-the-art VSOD methods on ASOD60K dataset.
    }
    \label{fig:vis}
\end{figure*}

\subsection{Main Results}
In this section, we compare our method with previous state-of-the-art methods, including CPD-R \cite{wu2019cascaded}, MINet \cite{pang2020multi}, SCRN \cite{wu2019stacked}, F3Net \cite{wei2020f3net},  LDF \cite{wei2020label}, CSFR2 \cite{gao2020highly}, GateNet \cite{zhao2020suppress}, COSNet \cite{lu2019see}, RCRNet \cite{yan2019semi}, PCSA \cite{gu2020pyramid}, 3DC-Seg \cite{mahadevan2020making} and RTNet \cite{ren2021reciprocal} on ASOD60K dataset.
\paragraph{Quantitative results.} We compare our method with state-of-the-art methods on the ASOD60K dataset in Table~\ref{tab:main results}. In general, our method achieves the best result of 0.404 $F_\beta$, 0.678 $S_\alpha$, 0.732 $E_\phi$ and 0.026 $\mathcal{M}$ on the ASOD60K test set. For each sound event split, all metrics of our method eclipse other methods on both Music and Speaking splits. While the 0.617 $E_\phi$ of our method on the Miscellanea split is slightly lower than the 0.644 $E_\phi$ of RCRNet \cite{yan2019semi}. Two reasons maybe account for the inferior performance of our method. First, the audio recordings in Miscellanea split contain background music which barriers our model to accurately locating the sound sources. Second, there are several unseen sound event classes in the Miscellanea test split. In this way, it is difficult for the model to construct correct multimodal correspondence between videos and audios with unseen classes. In the music and speaking scenario where sound sources can be easily localized, our method achieves obvious improvement against previous methods. 

\begin{figure}[h!]
    \centering
    \includegraphics[width=\linewidth]{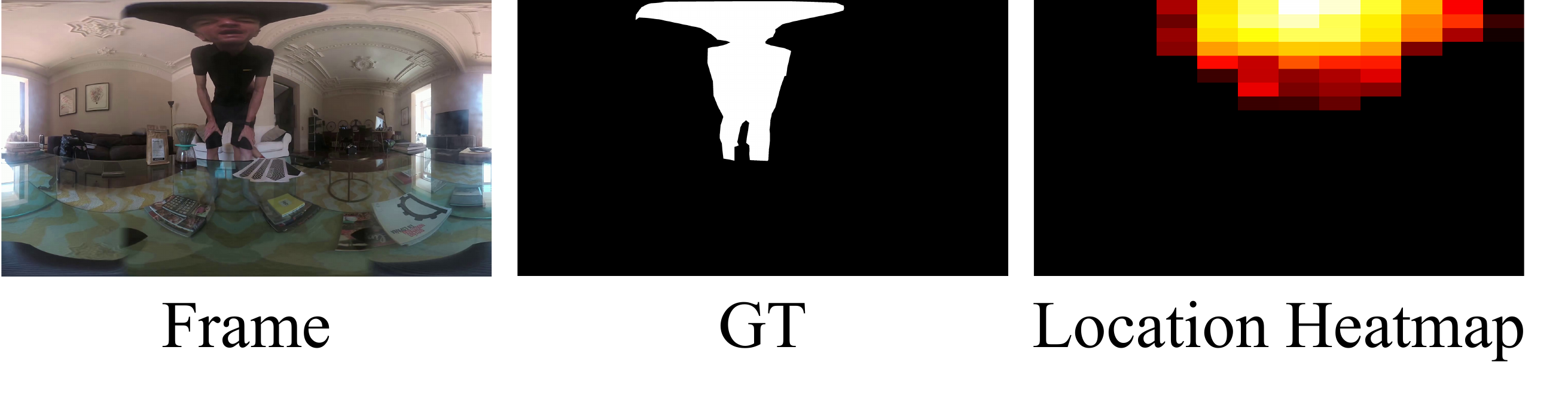}
    \vspace{-0.6cm}
    \caption{Visualization of sound source localization heatmap.
    }
    \label{fig:loc}
\end{figure}

\paragraph{Qualitative results.}
We present our qualitative result in Figure~\ref{fig:vis} and compare it against previous 2D methods on ASOD60K dataset. The result indicates that previous methods fail to detect the correct salient objects. In contrast, our method shows great accuracy and robustness even in very challenging scenarios, e.g., with severe distortions in the polar area as shown in the first line in Figure~\ref{fig:vis}. This implies that our network equipped with ACF block and SPE generates more accurate results than simply adopting previous 2D methods on the panoramic scenario.
We visualize the audio-guided location heatmap $\varphi_{C\mapsto 1}(f_t^{loc})$ in Figure~\ref{fig:loc}. We notice that the audio-guided location heatmap $\varphi_{C\mapsto 1}(f_t^{loc})$ reflects the correct location of sound sources thus helping the final salient object detection.

\begin{table}[t]
    \centering
\scalebox{1}{
    \begin{tabular}{l|p{0.5cm}<{\centering}p{0.5cm}<{\centering}p{0.5cm}<{\centering}p{0.5cm}<{\centering}} 
    \hline
    \hline
    \multirow{2}*{Multimodal Fusion} &   \multicolumn{4}{c}{\quad ASOD60K-Test All\quad}\\
    \cline{2-5}
    ~ & $F_\beta\uparrow$ & $S_\alpha\uparrow$ & $E_\phi\uparrow$ & $\mathcal{M}\downarrow$\\
    \hline
    None & .396 & .660 & .714 & .037\\
    +ACF (Concat) & .385 & .667 & .697 & .027 \\
    +ACF (MM Attn) & .397 & .670 & .722 & \bf.026 \\
    +ACF (MM Attn)+SPE & \bf.404 & \bf.678 & \bf.732 & \bf.026 \\
    \hline
    \hline
    \end{tabular}
}
\caption{\textbf{Impact of different multimodal fusion methods.} The content in the bracket indicates different fusion methods in ACF block. Concat: Concatenate, MM Attn: multimodal attention, SPE: spherical positional encoding. }
\label{tab:mm fusion}
\end{table}

\begin{table*}[t]
\begin{minipage}[t]{\textwidth}
\begin{minipage}[t]{0.47\textwidth}
\makeatletter\def\@captype{table}
\centering
\scalebox{1}{
    \begin{tabular}{l|p{0.5cm}<{\centering}p{0.5cm}<{\centering}p{0.5cm}<{\centering}p{0.5cm}<{\centering}} 
    \hline
    \hline
    \multirow{2}*{Backbone} &   \multicolumn{4}{c}{\quad ASOD60K-Test All\quad}\\
    \cline{2-5}
    ~ & $F_\beta\uparrow$ & $S_\alpha\uparrow$ & $E_\phi\uparrow$ & $\mathcal{M}\downarrow$\\
    \hline
    Backbone & .396 & .660 & .714 & .037\\
    +Transformer & \bf.404 & \bf.678 & .732 & \bf.026 \\
    +Transformer+SPE & .403 & .676 & \bf.742 & \bf.026 \\
    \hline
    \hline
    \end{tabular}
}
\caption{\textbf{Impact of visual feature extraction methods.} Components are added step by step.}
\label{tab:encoder}
\end{minipage}
\hspace{0.04\textwidth}
\begin{minipage}[t]{0.47\textwidth}
\makeatletter\def\@captype{table}
\centering
\scalebox{1}{
    \begin{tabular}{l|p{0.5cm}<{\centering}p{0.5cm}<{\centering}p{0.5cm}<{\centering}p{0.5cm}<{\centering}} 
    \hline
    \hline
    \multirow{2}*{$\lambda_{distill}$} &   \multicolumn{4}{c}{\quad ASOD60K-Test All\quad}\\
    \cline{2-5}
    ~ & $F_\beta\uparrow$ & $S_\alpha\uparrow$ & $E_\phi\uparrow$ & $\mathcal{M}\downarrow$\\
    \hline
    0.0 & .389 & .667 & .716 & .027\\
    1.0 & .402 & .676 & .725 & .029 \\
    2.0 & \bf.410 & .670 & .715 & .035 \\
    5.0 & .404 & \bf.678 & \bf.732 & \bf.026 \\
    \hline
    \hline
    \end{tabular}
}
\caption{\textbf{Impact of distillation loss weight.}}
\label{tab:loss}
\end{minipage}
\end{minipage}
\vspace{-0.2cm}
\end{table*}

\begin{table*}[t]
\begin{minipage}[t]{\textwidth}
\begin{minipage}[t]{0.47\textwidth}
\makeatletter\def\@captype{table}
\centering
\scalebox{1}{
    \begin{tabular}{c|p{0.5cm}<{\centering}p{0.5cm}<{\centering}p{0.5cm}<{\centering}p{0.5cm}<{\centering}} 
    \hline
    \hline
    \multirow{2}*{3D Localization} &   \multicolumn{4}{c}{\quad ASOD60K-Test All\quad}\\
    \cline{2-5}
    ~ & $F_\beta\uparrow$ & $S_\alpha\uparrow$ & $E_\phi\uparrow$ & $\mathcal{M}\downarrow$\\
    \hline
    \XSolidBrush & .391 & .667 & .723 & .030 \\
    \Checkmark & \bf.404 & \bf.678 & \bf.732 & \bf.026 \\
    \hline
    \hline
    \end{tabular}
}
\caption{\textbf{Impact of 3D sound source localization branch.}}
\label{tab:localization}
\end{minipage}
\hspace{0.04\textwidth}
\begin{minipage}[t]{0.47\textwidth}
\makeatletter\def\@captype{table}
\centering
\scalebox{1}{
    \begin{tabular}{c|p{0.5cm}<{\centering}p{0.5cm}<{\centering}p{0.5cm}<{\centering}p{0.5cm}<{\centering}} 
    \hline
    \hline
    \multirow{2}*{Window Size} &   \multicolumn{4}{c}{\quad ASOD60K-Test All\quad}\\
    \cline{2-5}
    ~ & $F_\beta\uparrow$ & $S_\alpha\uparrow$ & $E_\phi\uparrow$ & $\mathcal{M}\downarrow$\\
    \hline
    1 & .392 & .670 & .725 & .026 \\
    2 & .396 & .672 & .719 & .027 \\
    3 & \bf.404 & .678 & \bf.732 & \bf.026 \\
    4 & .402 & \bf.680 & .728 & .028 \\
    \hline
    \hline
    \end{tabular}
}
\caption{\textbf{Impact of input window size.}}
\label{tab:window}
\end{minipage}
\end{minipage}
\vspace{-0.2cm}
\end{table*}

\begin{table}[h]
    \centering
    \begin{tabular}{c|p{0.5cm}<{\centering}p{0.5cm}<{\centering}p{0.5cm}<{\centering}p{0.5cm}<{\centering}} 
    \hline
    \hline
    \multirow{2}*{Sound Type} & \multicolumn{4}{c}{\quad ASOD60K-Test All\quad}\\
    \cline{2-5}
    ~ & $F_\beta\uparrow$ & $S_\alpha\uparrow$ & $E_\phi\uparrow$ & $\mathcal{M}\downarrow$\\
    \hline
    None & .396 & .660 & .714 & .037 \\
    Mono & .397 & .662 & .717 & .029 \\
    Ambisonic & \bf.404 & \bf.678 & \bf.732 & \bf.026 \\
    \hline
    \hline
    \end{tabular}
\caption{\textbf{Impact of sound type.}}
\label{tab:sound}
\vspace{-0.2cm}
\end{table}

\subsection{Ablation Experiments}
We conduct extensive ablation studies on the ASOD60K dataset to verify the effectiveness of different components.

\paragraph{Multimodal fusion method.}
To investigate the effectiveness of our proposed ACF block, we conduct experiments with different multimodal fusion schemes. As shown in Table~\ref{tab:mm fusion}, we compare the ACF block equipped with multimodal attention with two baseline settings.  `None' and `ACF (Concat)' means no multimodal fusion and simply concatenating visual and acoustic features in the ACF block respectively. In general, the model without multimodal fusion leads to an inferior performance which indicates the acoustic modality is essential to salient object detection. We notice that ACF block equipped with multimodal attention outperforms the ACF block with simply multimodal feature concatenation. Additional spherical positional encoding (SPE) brings another 0.07 $F_{\beta}$, 0.08 $S_{\alpha}$ and 0.1 $E_{\phi}$ gain compared to multimodal attention.

\paragraph{Visual feature extraction.} We conduct experiments to ablate the influence of different visual feature extraction methods. We first build a baseline model that leverages ResNet-50 \cite{he2016deep} backbone to extract visual features which leads to 0.396 $F_{\beta}$, 0.660 $S_{\alpha}$, 0.714 $E_{\phi}$ and 0.37 $\mathcal{M}$. By employing a transformer encoder on top of the backbone, we observe non-trivial gains on all metrics. We consider the improvement comes from the strong global understanding capability of transformer. By replacing the standard positional encoding in the transformer with our spherical positional encoder, the $E_\phi$ metric improves 0.1 while $F_\beta$ and $S_\alpha$ slightly drop. We consider the performance drop mainly because the pretraining is conducted on the 2D dataset.

\paragraph{Distillation loss weight.} To investigate the influence of distillation loss weight, we conduct experiments by ablating different $\lambda_{distill}$. As shown in Table~\ref{tab:loss}, we notice that a weight of 5 leads to the best result of 0.404 $F_{\beta}$, 0.678 $S_{\alpha}$, 0.732 $E_{\phi}$ and 0.26 $\mathcal{M}$. The $\lambda_{distill}=0$ means that no teacher branch is adopted which leads to the worst result.

\paragraph{3D sound source localization.} To demonstrate the effectiveness of employing 3D sound source localization in VSOD, we conduct an experiment to disable the sound source localization branch in our method. The result in Table~\ref{tab:localization} indicates that 3D sound source localization from ambisonic audio can help salient object detection in panoramic scenarios.

\paragraph{Window size.} 
Since temporal information is essential for the video salient object detection, we conduct experiments on different input window sizes as shown in Table~\ref{tab:window}. We notice that the window size of 3 achieves the best performance in terms of $F_\beta$, $E_\phi$ and $\mathcal{M}$ and the window size of 4 achieves the best result in terms of $S_\alpha$.

\paragraph{Sound type.}
We conduct experiments to show the benefit of utilizing ambisonic audios. Table~\ref{tab:sound} shows that mono audio only has a trivial improvement compared to the baseline setting (no multimodal fusion). We consider this is because mono audio cannot explicitly model spatial information of the sound source.
\section{Conclusion}
In this paper, we propose a framework for audio-visual video salient object detection in panoramic scenarios. In particular, we propose an audio-visual context fusion block to enhance visual features by ambisonic audios. To better utilize the spatial information encoded in the audios, a label-guided distillation scheme is introduced to help the multimodal interaction. In addition, due to the severe distortions in the ER frame, we leverage position-agnostic attention mechanism equipped with spherical positional encoding to map each pixel back to 3D space thus capturing the true spatial location of each pixel. Notably, our method achieves the best result on the ASOD60K benchmark. Moreover, extensive study shows that ambisonic audio can help the salient object detection in panoramic videos.

\newpage
\clearpage
{\small
\bibliography{aaai22.bib}

\begin{thebibliography}{67}
\providecommand{\natexlab}[1]{#1}

\bibitem[{Achanta et~al.(2009)Achanta, Hemami, Estrada, and
  Susstrunk}]{achanta2009frequency}
Achanta, R.; Hemami, S.; Estrada, F.; and Susstrunk, S. 2009.
\newblock Frequency-tuned salient region detection.
\newblock In \emph{2009 IEEE conference on computer vision and pattern
  recognition}, 1597--1604. IEEE.

\bibitem[{Adavanne et~al.(2018)Adavanne, Politis, Nikunen, and
  Virtanen}]{adavanne2018sound}
Adavanne, S.; Politis, A.; Nikunen, J.; and Virtanen, T. 2018.
\newblock Sound event localization and detection of overlapping sources using
  convolutional recurrent neural networks.
\newblock \emph{IEEE Journal of Selected Topics in Signal Processing}, 13(1):
  34--48.

\bibitem[{Bertasius et~al.(2017)Bertasius, Soo~Park, Yu, and
  Shi}]{bertasius2017unsupervised}
Bertasius, G.; Soo~Park, H.; Yu, S.~X.; and Shi, J. 2017.
\newblock Unsupervised learning of important objects from first-person videos.
\newblock In \emph{Proceedings of the IEEE International Conference on Computer
  Vision}, 1956--1964.

\bibitem[{Borji et~al.(2015)Borji, Cheng, Jiang, and Li}]{borji2015salient}
Borji, A.; Cheng, M.-M.; Jiang, H.; and Li, J. 2015.
\newblock Salient object detection: A benchmark.
\newblock \emph{IEEE transactions on image processing}, 24(12): 5706--5722.

\bibitem[{Cai et~al.(2022)Cai, Li, Wang, and Wang}]{cai2022overview}
Cai, Y.; Li, X.; Wang, Y.; and Wang, R. 2022.
\newblock An Overview of Panoramic Video Projection Schemes in the IEEE 1857.9
  Standard for Immersive Visual Content Coding.
\newblock \emph{IEEE Transactions on Circuits and Systems for Video
  Technology}.

\bibitem[{Chao et~al.(2020)Chao, Ozcinar, Wang, Zerman, Zhang, Hamidouche,
  Deforges, and Smolic}]{chao2020audio}
Chao, F.-Y.; Ozcinar, C.; Wang, C.; Zerman, E.; Zhang, L.; Hamidouche, W.;
  Deforges, O.; and Smolic, A. 2020.
\newblock Audio-visual perception of omnidirectional video for virtual reality
  applications.
\newblock In \emph{2020 IEEE International Conference on Multimedia \& Expo
  Workshops (ICMEW)}, 1--6. IEEE.

\bibitem[{Chen et~al.(2022)Chen, Jin, Shen, and Yang}]{chen2022video}
Chen, Y.-W.; Jin, X.; Shen, X.; and Yang, M.-H. 2022.
\newblock Video Salient Object Detection via Contrastive Features and Attention
  Modules.
\newblock In \emph{Proceedings of the IEEE/CVF Winter Conference on
  Applications of Computer Vision}, 1320--1329.

\bibitem[{Cheng et~al.(2018)Cheng, Chao, Dong, Wen, Liu, and
  Sun}]{cheng2018cube}
Cheng, H.-T.; Chao, C.-H.; Dong, J.-D.; Wen, H.-K.; Liu, T.-L.; and Sun, M.
  2018.
\newblock Cube padding for weakly-supervised saliency prediction in 360 videos.
\newblock In \emph{Proceedings of the IEEE Conference on Computer Vision and
  Pattern Recognition}, 1420--1429.

\bibitem[{Cheng et~al.(2021)Cheng, Song, Tang, and Guo}]{cheng2021audio}
Cheng, S.; Song, L.; Tang, J.; and Guo, S. 2021.
\newblock Audio-Visual Salient Object Detection.
\newblock In \emph{International Conference on Intelligent Computing},
  510--521. Springer.

\bibitem[{Dosovitskiy et~al.(2020)Dosovitskiy, Beyer, Kolesnikov, Weissenborn,
  Zhai, Unterthiner, Dehghani, Minderer, Heigold, Gelly
  et~al.}]{dosovitskiy2020image}
Dosovitskiy, A.; Beyer, L.; Kolesnikov, A.; Weissenborn, D.; Zhai, X.;
  Unterthiner, T.; Dehghani, M.; Minderer, M.; Heigold, G.; Gelly, S.; et~al.
  2020.
\newblock An image is worth 16x16 words: Transformers for image recognition at
  scale.
\newblock \emph{arXiv preprint arXiv:2010.11929}.

\bibitem[{Fan et~al.(2017)Fan, Cheng, Liu, Li, and Borji}]{fan2017structure}
Fan, D.-P.; Cheng, M.-M.; Liu, Y.; Li, T.; and Borji, A. 2017.
\newblock Structure-measure: A new way to evaluate foreground maps.
\newblock In \emph{Proceedings of the IEEE international conference on computer
  vision}, 4548--4557.

\bibitem[{Fan et~al.(2018)Fan, Gong, Cao, Ren, Cheng, and
  Borji}]{fan2018enhanced}
Fan, D.-P.; Gong, C.; Cao, Y.; Ren, B.; Cheng, M.-M.; and Borji, A. 2018.
\newblock Enhanced-alignment measure for binary foreground map evaluation.
\newblock \emph{arXiv preprint arXiv:1805.10421}.

\bibitem[{Fan et~al.(2019)Fan, Wang, Cheng, and Shen}]{fan2019shifting}
Fan, D.-P.; Wang, W.; Cheng, M.-M.; and Shen, J. 2019.
\newblock Shifting more attention to video salient object detection.
\newblock In \emph{Proceedings of the IEEE/CVF conference on computer vision
  and pattern recognition}, 8554--8564.

\bibitem[{Fragkiadaki, Zhang, and Shi(2012)}]{fragkiadaki2012video}
Fragkiadaki, K.; Zhang, G.; and Shi, J. 2012.
\newblock Video segmentation by tracing discontinuities in a trajectory
  embedding.
\newblock In \emph{2012 IEEE Conference on Computer Vision and Pattern
  Recognition}, 1846--1853. IEEE.

\bibitem[{Gao et~al.(2020)Gao, Tan, Cheng, Lu, Chen, and Yan}]{gao2020highly}
Gao, S.-H.; Tan, Y.-Q.; Cheng, M.-M.; Lu, C.; Chen, Y.; and Yan, S. 2020.
\newblock Highly efficient salient object detection with 100k parameters.
\newblock In \emph{European Conference on Computer Vision}, 702--721. Springer.

\bibitem[{Gu et~al.(2020)Gu, Wang, Wang, Liu, Cheng, and Lu}]{gu2020pyramid}
Gu, Y.; Wang, L.; Wang, Z.; Liu, Y.; Cheng, M.-M.; and Lu, S.-P. 2020.
\newblock Pyramid constrained self-attention network for fast video salient
  object detection.
\newblock In \emph{Proceedings of the AAAI conference on artificial
  intelligence}, volume~34, 10869--10876.

\bibitem[{Guizzo et~al.(2021)Guizzo, Gramaccioni, Jamili, Marinoni, Massaro,
  Medaglia, Nachira, Nucciarelli, Paglialunga, Pennese
  et~al.}]{guizzo2021l3das21}
Guizzo, E.; Gramaccioni, R.~F.; Jamili, S.; Marinoni, C.; Massaro, E.;
  Medaglia, C.; Nachira, G.; Nucciarelli, L.; Paglialunga, L.; Pennese, M.;
  et~al. 2021.
\newblock L3DAS21 Challenge: Machine learning for 3D audio signal processing.
\newblock In \emph{2021 IEEE 31st International Workshop on Machine Learning
  for Signal Processing (MLSP)}, 1--6. IEEE.

\bibitem[{He et~al.(2016)He, Zhang, Ren, and Sun}]{he2016deep}
He, K.; Zhang, X.; Ren, S.; and Sun, J. 2016.
\newblock Deep residual learning for image recognition.
\newblock In \emph{Proceedings of the IEEE conference on computer vision and
  pattern recognition}, 770--778.

\bibitem[{Huang et~al.(2021{\natexlab{a}})Huang, Li, Wang, Jiang, and
  Zhang}]{huang2021forgery}
Huang, Y.; Li, X.; Wang, W.; Jiang, T.; and Zhang, Q. 2021{\natexlab{a}}.
\newblock Forgery Attack Detection in Surveillance Video Streams Using Wi-Fi
  Channel State Information.
\newblock \emph{IEEE Transactions on Wireless Communications}.

\bibitem[{Huang et~al.(2021{\natexlab{b}})Huang, Li, Wang, Jiang, and
  Zhang}]{huang2021towards}
Huang, Y.; Li, X.; Wang, W.; Jiang, T.; and Zhang, Q. 2021{\natexlab{b}}.
\newblock Towards cross-modal forgery detection and localization on live
  surveillance videos.
\newblock In \emph{IEEE INFOCOM 2021-IEEE Conference on Computer
  Communications}, 1--10. IEEE.

\bibitem[{Jiang and Davis(2013)}]{jiang2013submodular}
Jiang, Z.; and Davis, L.~S. 2013.
\newblock Submodular salient region detection.
\newblock In \emph{Proceedings of the IEEE conference on computer vision and
  pattern recognition}, 2043--2050.

\bibitem[{Kim and Hwang(2002)}]{kim2002fast}
Kim, C.; and Hwang, J.-N. 2002.
\newblock Fast and automatic video object segmentation and tracking for
  content-based applications.
\newblock \emph{IEEE transactions on circuits and systems for video
  technology}, 12(2): 122--129.

\bibitem[{Le and Sugimoto(2017)}]{le2017deeply}
Le, T.-N.; and Sugimoto, A. 2017.
\newblock Deeply Supervised 3D Recurrent FCN for Salient Object Detection in
  Videos.
\newblock In \emph{BMVC}, volume~1, 3.

\bibitem[{Li et~al.(2018)Li, Xie, Wei, Wang, and Lin}]{li2018flow}
Li, G.; Xie, Y.; Wei, T.; Wang, K.; and Lin, L. 2018.
\newblock Flow guided recurrent neural encoder for video salient object
  detection.
\newblock In \emph{Proceedings of the IEEE conference on computer vision and
  pattern recognition}, 3243--3252.

\bibitem[{Li et~al.(2019)Li, Chen, Li, and Yu}]{li2019motion}
Li, H.; Chen, G.; Li, G.; and Yu, Y. 2019.
\newblock Motion guided attention for video salient object detection.
\newblock In \emph{Proceedings of the IEEE/CVF international conference on
  computer vision}, 7274--7283.

\bibitem[{Li et~al.(2022{\natexlab{a}})Li, Wang, Li, and Lu}]{li2022hybrid}
Li, X.; Wang, J.; Li, X.; and Lu, Y. 2022{\natexlab{a}}.
\newblock Hybrid instance-aware temporal fusion for online video instance
  segmentation.
\newblock In \emph{Proceedings of the AAAI Conference on Artificial
  Intelligence}, volume~36, 1429--1437.

\bibitem[{Li et~al.(2022{\natexlab{b}})Li, Wang, Li, and Lu}]{li2022video}
Li, X.; Wang, J.; Li, X.; and Lu, Y. 2022{\natexlab{b}}.
\newblock Video instance segmentation by instance flow assembly.
\newblock \emph{IEEE Transactions on Multimedia}.

\bibitem[{Li et~al.(2022{\natexlab{c}})Li, Wang, Xu, Li, Lu, and Raj}]{li2022r}
Li, X.; Wang, J.; Xu, X.; Li, X.; Lu, Y.; and Raj, B. 2022{\natexlab{c}}.
\newblock R\^{} 2VOS: Robust Referring Video Object Segmentation via Relational
  Multimodal Cycle Consistency.
\newblock \emph{arXiv preprint arXiv:2207.01203}.

\bibitem[{Li et~al.(2022{\natexlab{d}})Li, Wang, Xu, Raj, and
  Lu}]{li2022online}
Li, X.; Wang, J.; Xu, X.; Raj, B.; and Lu, Y. 2022{\natexlab{d}}.
\newblock Online Video Instance Segmentation via Robust Context Fusion.
\newblock \emph{arXiv preprint arXiv:2207.05580}.

\bibitem[{Lin et~al.(2017)Lin, Doll{\'a}r, Girshick, He, Hariharan, and
  Belongie}]{lin2017feature}
Lin, T.-Y.; Doll{\'a}r, P.; Girshick, R.; He, K.; Hariharan, B.; and Belongie,
  S. 2017.
\newblock Feature pyramid networks for object detection.
\newblock In \emph{Proceedings of the IEEE conference on computer vision and
  pattern recognition}, 2117--2125.

\bibitem[{Liu et~al.(2021)Liu, Tan, He, and Xiao}]{liu2021swinnet}
Liu, Z.; Tan, Y.; He, Q.; and Xiao, Y. 2021.
\newblock SwinNet: Swin Transformer drives edge-aware RGB-D and RGB-T salient
  object detection.
\newblock \emph{IEEE Transactions on Circuits and Systems for Video
  Technology}.

\bibitem[{Loshchilov and Hutter(2017)}]{loshchilov2017decoupled}
Loshchilov, I.; and Hutter, F. 2017.
\newblock Decoupled weight decay regularization.
\newblock \emph{arXiv preprint arXiv:1711.05101}.

\bibitem[{Lu et~al.(2019)Lu, Wang, Ma, Shen, Shao, and Porikli}]{lu2019see}
Lu, X.; Wang, W.; Ma, C.; Shen, J.; Shao, L.; and Porikli, F. 2019.
\newblock See more, know more: Unsupervised video object segmentation with
  co-attention siamese networks.
\newblock In \emph{Proceedings of the IEEE/CVF conference on computer vision
  and pattern recognition}, 3623--3632.

\bibitem[{Mahadevan et~al.(2020)Mahadevan, Athar, O{\v{s}}ep, Hennen,
  Leal-Taix{\'e}, and Leibe}]{mahadevan2020making}
Mahadevan, S.; Athar, A.; O{\v{s}}ep, A.; Hennen, S.; Leal-Taix{\'e}, L.; and
  Leibe, B. 2020.
\newblock Making a case for 3d convolutions for object segmentation in videos.
\newblock \emph{arXiv preprint arXiv:2008.11516}.

\bibitem[{Milletari, Navab, and Ahmadi(2016)}]{milletari2016v}
Milletari, F.; Navab, N.; and Ahmadi, S.-A. 2016.
\newblock V-net: Fully convolutional neural networks for volumetric medical
  image segmentation.
\newblock In \emph{2016 fourth international conference on 3D vision (3DV)},
  565--571. IEEE.

\bibitem[{Nguyen, Yan, and Nahrstedt(2018)}]{nguyen2018your}
Nguyen, A.; Yan, Z.; and Nahrstedt, K. 2018.
\newblock Your attention is unique: Detecting 360-degree video saliency in
  head-mounted display for head movement prediction.
\newblock In \emph{Proceedings of the 26th ACM international conference on
  Multimedia}, 1190--1198.

\bibitem[{Pang et~al.(2020)Pang, Zhao, Zhang, and Lu}]{pang2020multi}
Pang, Y.; Zhao, X.; Zhang, L.; and Lu, H. 2020.
\newblock Multi-scale interactive network for salient object detection.
\newblock In \emph{Proceedings of the IEEE/CVF conference on computer vision
  and pattern recognition}, 9413--9422.

\bibitem[{Ren et~al.(2021{\natexlab{a}})Ren, Liu, Liu, Chen, Han, and
  He}]{ren2021reciprocal}
Ren, S.; Liu, W.; Liu, Y.; Chen, H.; Han, G.; and He, S. 2021{\natexlab{a}}.
\newblock Reciprocal transformations for unsupervised video object
  segmentation.
\newblock In \emph{Proceedings of the IEEE/CVF conference on computer vision
  and pattern recognition}, 15455--15464.

\bibitem[{Ren et~al.(2021{\natexlab{b}})Ren, Wen, Zhao, Han, and
  He}]{ren2021unifying}
Ren, S.; Wen, Q.; Zhao, N.; Han, G.; and He, S. 2021{\natexlab{b}}.
\newblock Unifying Global-Local Representations in Salient Object Detection
  with Transformer.
\newblock \emph{arXiv preprint arXiv:2108.02759}.

\bibitem[{Song et~al.(2018)Song, Wang, Zhao, Shen, and Lam}]{song2018pyramid}
Song, H.; Wang, W.; Zhao, S.; Shen, J.; and Lam, K.-M. 2018.
\newblock Pyramid dilated deeper convlstm for video salient object detection.
\newblock In \emph{Proceedings of the European conference on computer vision
  (ECCV)}, 715--731.

\bibitem[{Su et~al.(2022)Su, Deng, Sun, Lin, and Wu}]{su2022unified}
Su, Y.; Deng, J.; Sun, R.; Lin, G.; and Wu, Q. 2022.
\newblock A Unified Transformer Framework for Group-based Segmentation:
  Co-Segmentation, Co-Saliency Detection and Video Salient Object Detection.
\newblock \emph{arXiv preprint arXiv:2203.04708}.

\bibitem[{Suzuki and Yamanaka(2018)}]{suzuki2018saliency}
Suzuki, T.; and Yamanaka, T. 2018.
\newblock Saliency map estimation for omni-directional image considering prior
  distributions.
\newblock In \emph{2018 IEEE International Conference on Systems, Man, and
  Cybernetics (SMC)}, 2079--2084. IEEE.

\bibitem[{Tsiami, Koutras, and Maragos(2020)}]{tsiami2020stavis}
Tsiami, A.; Koutras, P.; and Maragos, P. 2020.
\newblock Stavis: Spatio-temporal audiovisual saliency network.
\newblock In \emph{Proceedings of the IEEE/CVF Conference on Computer Vision
  and Pattern Recognition}, 4766--4776.

\bibitem[{Vaswani et~al.(2017)Vaswani, Shazeer, Parmar, Uszkoreit, Jones,
  Gomez, Kaiser, and Polosukhin}]{vaswani2017attention}
Vaswani, A.; Shazeer, N.; Parmar, N.; Uszkoreit, J.; Jones, L.; Gomez, A.~N.;
  Kaiser, {\L}.; and Polosukhin, I. 2017.
\newblock Attention is all you need.
\newblock \emph{Advances in neural information processing systems}, 30.

\bibitem[{Wang et~al.(2021)Wang, Jiang, Ren, Hu, and Bai}]{wang2021swiftnet}
Wang, H.; Jiang, X.; Ren, H.; Hu, Y.; and Bai, S. 2021.
\newblock Swiftnet: Real-time video object segmentation.
\newblock In \emph{Proceedings of the IEEE/CVF Conference on Computer Vision
  and Pattern Recognition}, 1296--1305.

\bibitem[{Wang et~al.(2017)Wang, Lu, Wang, Feng, Wang, Yin, and
  Ruan}]{wang2017learning}
Wang, L.; Lu, H.; Wang, Y.; Feng, M.; Wang, D.; Yin, B.; and Ruan, X. 2017.
\newblock Learning to detect salient objects with image-level supervision.
\newblock In \emph{Proceedings of the IEEE conference on computer vision and
  pattern recognition}, 136--145.

\bibitem[{Wang et~al.(2015)Wang, Shen, Li, and Porikli}]{wang2015robust}
Wang, W.; Shen, J.; Li, X.; and Porikli, F. 2015.
\newblock Robust video object cosegmentation.
\newblock \emph{IEEE Transactions on Image Processing}, 24(10): 3137--3148.

\bibitem[{Wang, Shen, and Porikli(2015)}]{wang2015saliency}
Wang, W.; Shen, J.; and Porikli, F. 2015.
\newblock Saliency-aware geodesic video object segmentation.
\newblock In \emph{Proceedings of the IEEE conference on computer vision and
  pattern recognition}, 3395--3402.

\bibitem[{Wang, Shen, and Shao(2017)}]{wang2017video}
Wang, W.; Shen, J.; and Shao, L. 2017.
\newblock Video salient object detection via fully convolutional networks.
\newblock \emph{IEEE Transactions on Image Processing}, 27(1): 38--49.

\bibitem[{Wang et~al.(2019)Wang, Song, Zhao, Shen, Zhao, Hoi, and
  Ling}]{wang2019learning}
Wang, W.; Song, H.; Zhao, S.; Shen, J.; Zhao, S.; Hoi, S.~C.; and Ling, H.
  2019.
\newblock Learning unsupervised video object segmentation through visual
  attention.
\newblock In \emph{Proceedings of the IEEE/CVF Conference on Computer Vision
  and Pattern Recognition}, 3064--3074.

\bibitem[{Wei, Wang, and Huang(2020)}]{wei2020f3net}
Wei, J.; Wang, S.; and Huang, Q. 2020.
\newblock F$^3$Net: fusion, feedback and focus for salient object detection.
\newblock In \emph{Proceedings of the AAAI Conference on Artificial
  Intelligence}, volume~34, 12321--12328.

\bibitem[{Wei et~al.(2020)Wei, Wang, Wu, Su, Huang, and Tian}]{wei2020label}
Wei, J.; Wang, S.; Wu, Z.; Su, C.; Huang, Q.; and Tian, Q. 2020.
\newblock Label decoupling framework for salient object detection.
\newblock In \emph{Proceedings of the IEEE/CVF conference on computer vision
  and pattern recognition}, 13025--13034.

\bibitem[{Wu et~al.(2022)Wu, Jiang, Sun, Yuan, and Luo}]{wu2022language}
Wu, J.; Jiang, Y.; Sun, P.; Yuan, Z.; and Luo, P. 2022.
\newblock Language as Queries for Referring Video Object Segmentation.
\newblock In \emph{Proceedings of the IEEE/CVF Conference on Computer Vision
  and Pattern Recognition}, 4974--4984.

\bibitem[{Wu, Su, and Huang(2019{\natexlab{a}})}]{wu2019cascaded}
Wu, Z.; Su, L.; and Huang, Q. 2019{\natexlab{a}}.
\newblock Cascaded partial decoder for fast and accurate salient object
  detection.
\newblock In \emph{Proceedings of the IEEE/CVF conference on computer vision
  and pattern recognition}, 3907--3916.

\bibitem[{Wu, Su, and Huang(2019{\natexlab{b}})}]{wu2019stacked}
Wu, Z.; Su, L.; and Huang, Q. 2019{\natexlab{b}}.
\newblock Stacked cross refinement network for edge-aware salient object
  detection.
\newblock In \emph{Proceedings of the IEEE/CVF international conference on
  computer vision}, 7264--7273.

\bibitem[{Yan et~al.(2019)Yan, Li, Xie, Li, Wang, Chen, and Lin}]{yan2019semi}
Yan, P.; Li, G.; Xie, Y.; Li, Z.; Wang, C.; Chen, T.; and Lin, L. 2019.
\newblock Semi-supervised video salient object detection using pseudo-labels.
\newblock In \emph{Proceedings of the IEEE/CVF international conference on
  computer vision}, 7284--7293.

\bibitem[{Yang et~al.(2013)Yang, Zhang, Lu, Ruan, and Yang}]{yang2013saliency}
Yang, C.; Zhang, L.; Lu, H.; Ruan, X.; and Yang, M.-H. 2013.
\newblock Saliency detection via graph-based manifold ranking.
\newblock In \emph{Proceedings of the IEEE conference on computer vision and
  pattern recognition}, 3166--3173.

\bibitem[{Ye et~al.(2019)Ye, Rochan, Liu, and Wang}]{ye2019cross}
Ye, L.; Rochan, M.; Liu, Z.; and Wang, Y. 2019.
\newblock Cross-modal self-attention network for referring image segmentation.
\newblock In \emph{Proceedings of the IEEE/CVF conference on computer vision
  and pattern recognition}, 10502--10511.

\bibitem[{Zhang et~al.(2022)Zhang, Kang, Yang, Zhang, Zheng, and
  Sun}]{zhang2022lgd}
Zhang, P.; Kang, Z.; Yang, T.; Zhang, X.; Zheng, N.; and Sun, J. 2022.
\newblock LGD: Label-Guided Self-Distillation for Object Detection.
\newblock In \emph{Proceedings of the AAAI Conference on Artificial
  Intelligence}, volume~36, 3309--3317.

\bibitem[{Zhang, Chao, and Zhang(2021)}]{zhang2021asod60k}
Zhang, Y.; Chao, F.-Y.; and Zhang, L. 2021.
\newblock ASOD60K: An Audio-Induced Salient Object Detection Dataset for
  Panoramic Videos.
\newblock \emph{arXiv preprint arXiv:2107.11629}.

\bibitem[{Zhang et~al.(2018)Zhang, Xu, Yu, and Gao}]{zhang2018saliency}
Zhang, Z.; Xu, Y.; Yu, J.; and Gao, S. 2018.
\newblock Saliency detection in 360 videos.
\newblock In \emph{Proceedings of the European conference on computer vision
  (ECCV)}, 488--503.

\bibitem[{Zhang, Yu, and Crandall(2019)}]{zhang2019self}
Zhang, Z.; Yu, C.; and Crandall, D. 2019.
\newblock A self validation network for object-level human attention
  estimation.
\newblock \emph{Advances in Neural Information Processing Systems}, 32.

\bibitem[{Zhao et~al.(2022)Zhao, Li, Dong, and Younes}]{zhao2022self}
Zhao, C.; Li, X.; Dong, S.; and Younes, R. 2022.
\newblock Self-supervised Multi-Modal Video Forgery Attack Detection.
\newblock \emph{arXiv preprint arXiv:2209.06345}.

\bibitem[{Zhao et~al.(2020)Zhao, Pang, Zhang, Lu, and Zhang}]{zhao2020suppress}
Zhao, X.; Pang, Y.; Zhang, L.; Lu, H.; and Zhang, L. 2020.
\newblock Suppress and balance: A simple gated network for salient object
  detection.
\newblock In \emph{European conference on computer vision}, 35--51. Springer.

\bibitem[{Zhou et~al.(2022)Zhou, Wang, Zhang, Sun, Zhang, Birchfield, Guo,
  Kong, Wang, and Zhong}]{zhou2022audio}
Zhou, J.; Wang, J.; Zhang, J.; Sun, W.; Zhang, J.; Birchfield, S.; Guo, D.;
  Kong, L.; Wang, M.; and Zhong, Y. 2022.
\newblock Audio--Visual Segmentation.
\newblock In \emph{European Conference on Computer Vision}, 386--403. Springer.

\bibitem[{Zhou et~al.(2020)Zhou, Li, Wang, Tao, and Shen}]{zhou2020matnet}
Zhou, T.; Li, J.; Wang, S.; Tao, R.; and Shen, J. 2020.
\newblock Matnet: Motion-attentive transition network for zero-shot video
  object segmentation.
\newblock \emph{IEEE Transactions on Image Processing}, 29: 8326--8338.

\bibitem[{Zhu, Zhai, and Min(2018)}]{zhu2018prediction}
Zhu, Y.; Zhai, G.; and Min, X. 2018.
\newblock The prediction of head and eye movement for 360 degree images.
\newblock \emph{Signal Processing: Image Communication}, 69: 15--25.

\end{thebibliography}
}
\end{document}


\maketitle
\section{Audio-visual Correspondence}

\begin{figure}[h]
\centering
\includegraphics[width=\linewidth]{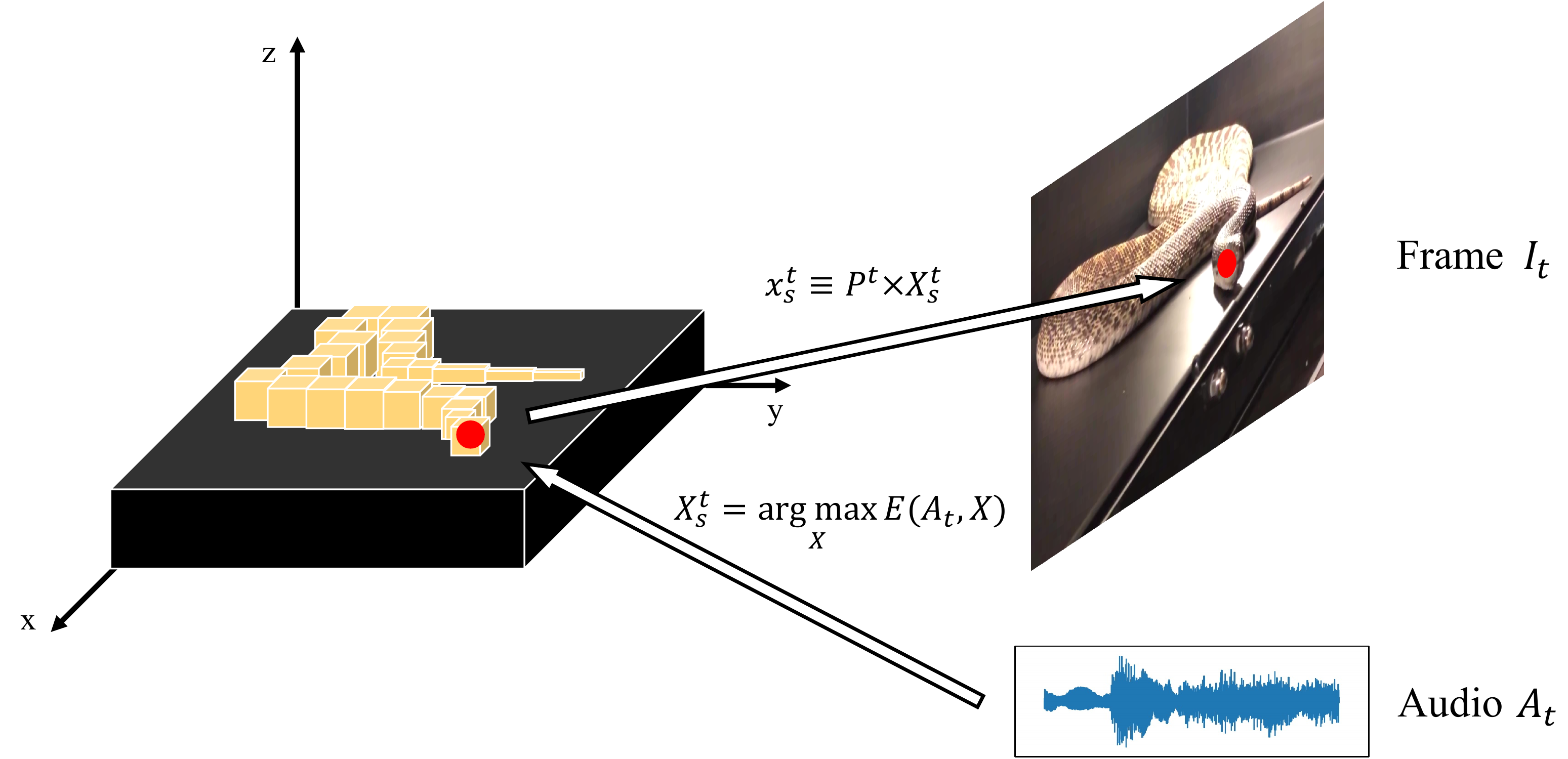}
\caption{\textbf{Audio-visual relationship.} }
\label{fig:av}
\end{figure}

Given the audio recordings $A^t$, we can decode the 3D location of sound source by signal processing methods such as MUSIC algorithm \cite{schmidt1986multiple} which searches for the location of the maximum of the spatial energy spectrum as
\begin{equation}
    X^t_s = \mathop{\arg \max}_{X} E(A^t, X)
\end{equation}
where $E(\cdot)$ is the spatial energy spectrum and $X^t_s\in\mathbb{R}^{3}$ is the 3D coordinate of sound source at time $t$. Therefore, the sound source can be easily corresponded to the pixel on a frame by applying camera matrix. Given that, we can represent the relationship between audio recordings $A^t$ and 2D location of sound source $x_s$ in frame $I_t$ by 
\begin{equation}
    x^t_s \equiv P^t\times \mathop{\arg \max}_{X} E(A^t, P)
\end{equation}
where $P^t$ is the camera matrix at time t and $x_t$ is the homogeneous coordinate of sound source in frame $I_t$. In practice, the camera matrix $P^t$ is unknown but can be estimated by a adjacent input frames $P^t=\mathcal{H}(I_t, I_{t-1})$. Consequently, we can link the sound source location $p^t_s$ in frames $I_t$ from audio $A_t$. 

\section{Detailed Structure of Acoustic Encoder}
The block diagram of the proposed method for SELD is presented in Figure~\ref{fig:crnn}. The input to the method is the multichannel audio. The phase and magnitude spectrograms are extracted from each audio channel and are used as separate features. The proposed method takes a sequence of features in consecutive spectrogram frames as input and predicts all the sound event classes active for each of the input frames along with their respective spatial location, producing the temporal activity and DOA trajectory for each sound event class. In particular, a CRNN is used to map the feature sequence to the two outputs in parallel. At the first output, SED is performed as a multi-label classification task, allowing the network to simultaneously estimate the presence of multiple sound events for each frame. At the second output, DOA estimates in the continuous 3D space are obtained as a multi-output regression task, where each sound event class is associated with three regressors that estimate the 3D Cartesian coordinates $x$, $y$ and $z$ of the DOA on a unit sphere around the microphone. The SED output of the network is in the continuous range of [0 1] for each sound event in the dataset, and this value is thresholded to obtain a binary decision for the respective sound event activity as shown in Figure~\ref{fig:crnn_output}. Finally, the respective DOA estimates for these active sound event classes provide their spatial locations. The detailed description of the feature extraction and the proposed method is explained in the following sections.

\subsection{Feature extraction}
\begin{figure}[htp]
  \centering  
     \subfigure[SELDnet]{{\includegraphics[height=16cm,width=\linewidth,keepaspectratio, trim=0.1cm 0.3cm 0.1cm 0.1cm,clip]{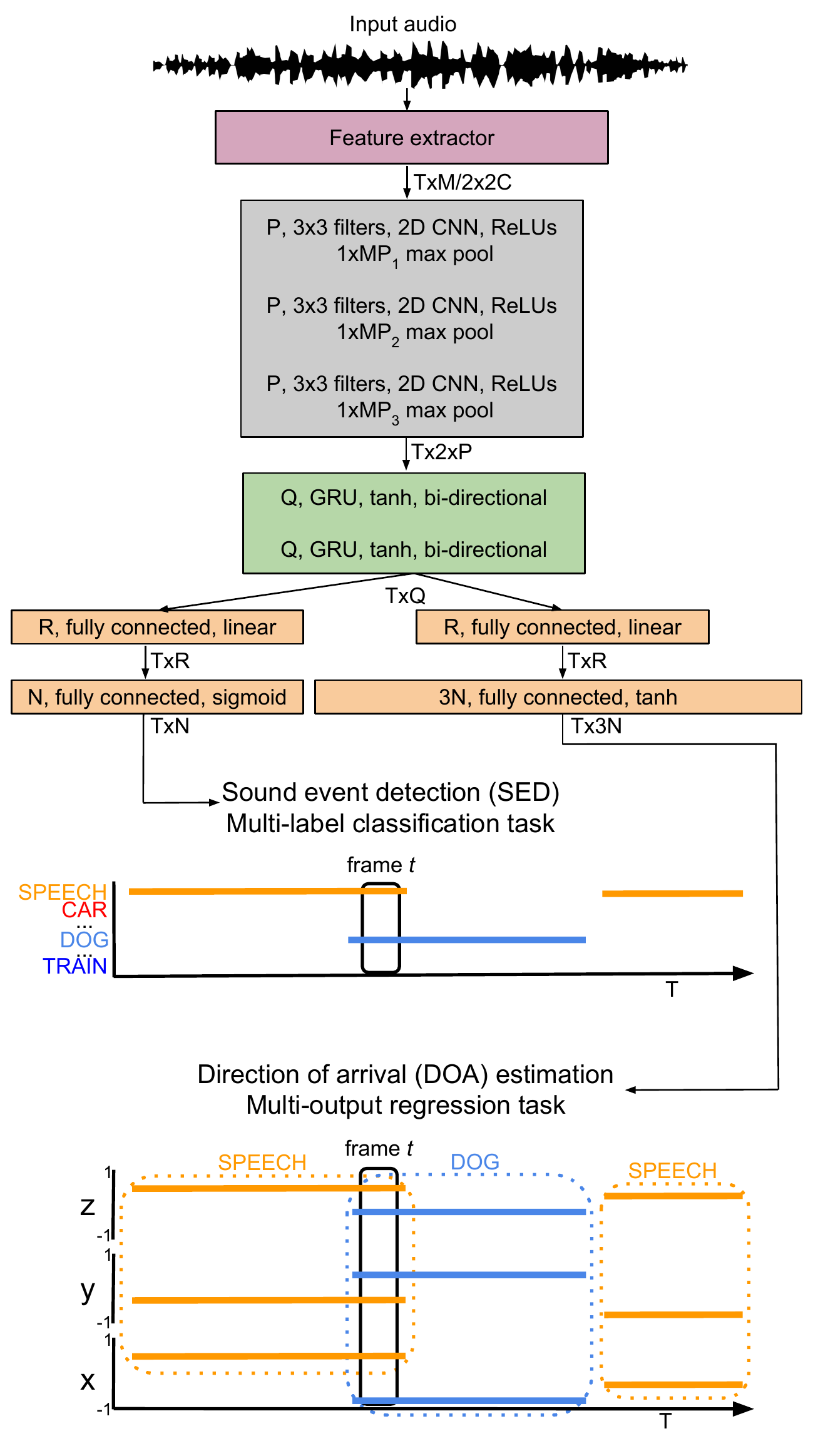} \label{fig:crnn}}}      
    \vspace{20pt}
    \subfigure[SELDnet output]{{\includegraphics[height=4.5cm, keepaspectratio, trim=0.3cm -0.2cm 0.5cm 0.1cm,clip]{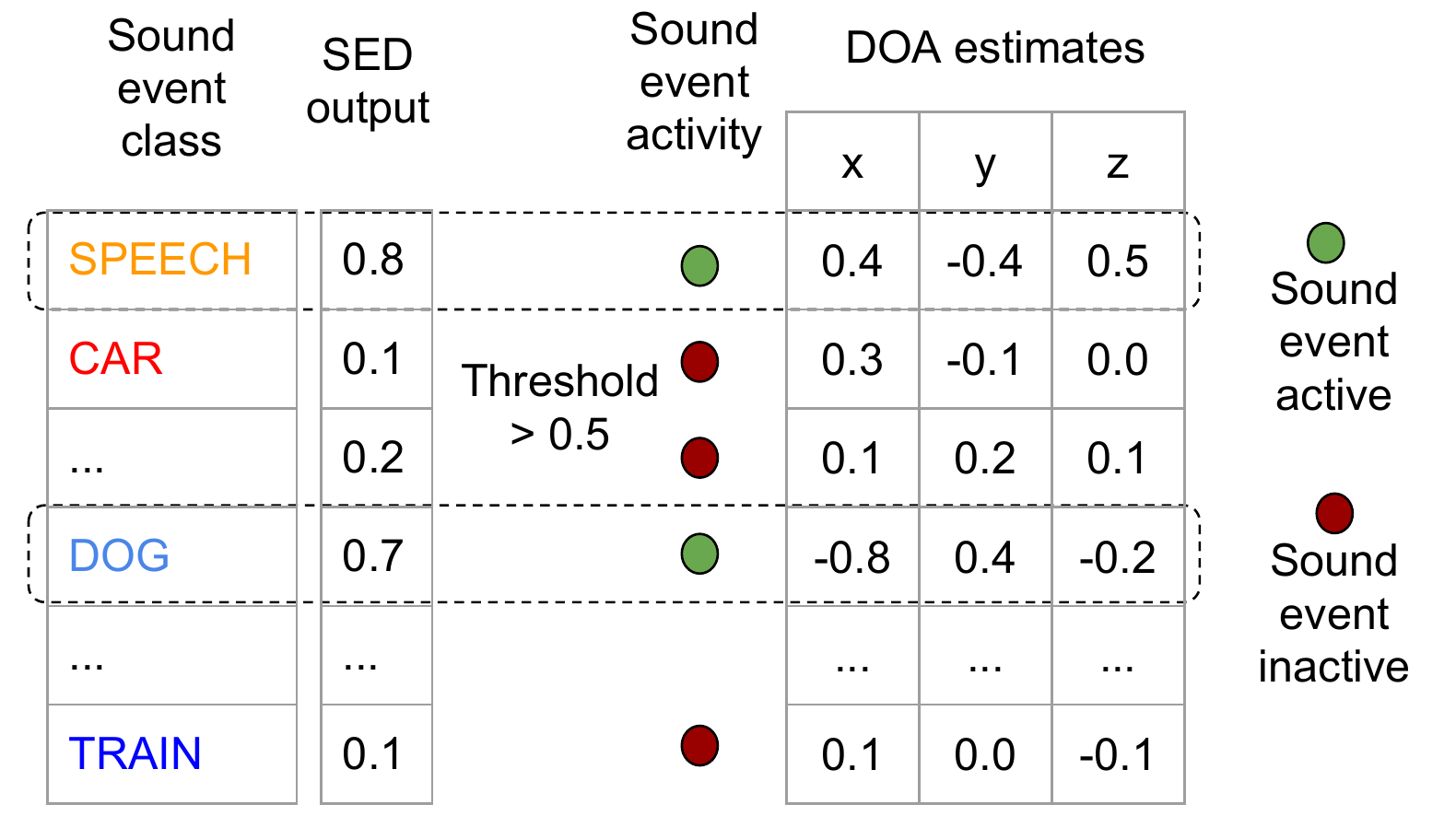} \label{fig:crnn_output}}}    
    \vspace{10pt}
  \caption{a) The proposed SELDnet and b) the frame-wise output for frame $t$ in Figure a). A sound event is said to be localized and detected when the confidence of the SED output exceeds the threshold.}
\end{figure}

The spectrogram is extracted from each of the $C$ channels of the multichannel audio using an $M$-point discrete Fourier transform (DFT) on Hamming window of length $M$ and 50\% overlap. The phase and magnitude of the spectrogram are then extracted and used as separate features. Only the $M/2$ positive frequencies without the zeroth bin are used. The output of the feature extraction block in Figure~\ref{fig:crnn} is a feature sequence of $T$ frames, with an overall dimension of $T\times M/2 \times 2C$, where the $2C$ dimension consists of $C$ magnitude and $C$ phase components.

\subsection{Neural network architecture}
The output of the feature extraction block is fed to the neural network as shown in Figure~\ref{fig:crnn}. In the proposed architecture the local shift-invariant features in the spectrogram are learned using multiple layers of 2D CNN. Each CNN layer has $P$ filters of $3\times3\times2C$ dimensional receptive fields acting along the time-frequency-channel axis with a rectified linear unit (ReLU) activation. The use of filter kernels spanning all the channels allows the CNN to learn relevant inter-channel features required for localization, whereas the time and frequency dimensions of the kernel allows learning relevant intra-channel features suitable for both the DOA and SED tasks. After each layer of CNN, the output activations are normalized using batch normalization~\cite{ioffe2015batch}, and the dimensionality is reduced using max-pooling ($MP_i$) along the frequency axis, thereby keeping the sequence length $T$ unchanged. The output after the final CNN layer with $P$ filters is of dimension $T\times2\times P$, where the reduced frequency dimension of $2$ is a result of max-pooling across CNN layers. 

The output activation from CNN is further reshaped to a $T$ frame sequence of length $2P$ feature vectors and fed to bidirectional RNN layers which are used to learn the temporal context information from the CNN output activations.  Specifically, $Q$ nodes of gated recurrent units (GRU) are used in each layer with tanh activations. This is followed by two branches of FC layers in parallel, one each for SED and DOA estimation. The FC layers share weights across time steps. The first FC layer in both the branches contains $R$ nodes each with linear activation. The last FC layer in the SED branch consists of $N$ nodes with sigmoid activation, each corresponding to one of the $N$ sound event classes to be detected. The use of sigmoid activation enables multiple classes to be active simultaneously. The last FC layer in the DOA branch consists of $3N$ nodes with tanh activation, where each of the $N$ sound event classes is represented by $3$ nodes corresponding to the sound event location in $x$, $y$, and $z$, respectively. For a DOA estimate on a unit sphere centered at the origin, the range of location along each axes is $[-1, 1]$, thus we use the tanh activation for these regressors to keep the output of the network in a similar range. 

We refer to the above architecture as SELDnet. The SED output of the SELDnet is in the continuous range of $[0, 1]$ for each class, while the DOA output is in the continuous range of $[-1, 1]$ for each axes of the sound class location. A sound event is said to be active, and its respective DOA estimate is chosen if the SED output exceeds the threshold of 0.5 as shown in Figure~\ref{fig:crnn_output}.

{\small
\bibliography{aaai22.bib}
}